\documentclass{article}

\PassOptionsToPackage{numbers, sort&compress}{natbib}

\usepackage[preprint]{neurips_2026}

\usepackage[utf8]{inputenc} 
\usepackage[T1]{fontenc}    
\usepackage{hyperref}       
\usepackage{url}            
\usepackage{booktabs}       
\usepackage{amsfonts}       
\usepackage{nicefrac}       
\usepackage{microtype}      
\usepackage{xcolor}         

\usepackage{graphicx}
\usepackage{subcaption}
\usepackage{nicematrix}
\usepackage{tikz}
\usepackage{diagbox}
\usepackage[skip=3pt]{caption}
\usepackage{wrapfig}
\usepackage{calc}
\usepackage{amsmath}
\usepackage{amssymb}
\usepackage{mathtools}
\usepackage{amsthm}
\usepackage[capitalize,noabbrev]{cleveref}
\usepackage[textsize=tiny]{todonotes}
\usepackage{enumitem}
\usepackage{floatflt}

\graphicspath{ {figures/} }

\newcommand{\indentedsay}[2][15pt]{
  \hspace*{#1}%
    \begin{minipage}{.8\textwidth}
       ``#2''
    \end{minipage}
}

\newcommand{\SplitCell}[4]{%
    \pgfmathtruncatemacro{\top}{#1}%
    \pgfmathtruncatemacro{\bottom}{#1+1}%
    \pgfmathtruncatemacro{\left}{#2}%
    \pgfmathtruncatemacro{\right}{#2+1}%
    \tikz \fill [#3] (\top-|\left) -- (\bottom-|\left) -- (\bottom-|\right) -- cycle;%
    \tikz \fill [#4] (\top-|\left) -- (\top-|\right) -- (\bottom-|\right) -- cycle;%
}

\theoremstyle{plain}

\theoremstyle{definition}

\theoremstyle{remark}

\title{Language Model Goal Selection Differs from Humans' in a Self-Directed Learning Task}

\author{%
  Gaia Molinaro \\
  University of California, Berkeley \& Amazon AGI Lab \\
  \texttt{gaiamolinaro@berkeley.edu} \\
  \And
  Dave August \\
  Amazon AGI Lab \\
  \And
  Danielle Perszyk \\
  Amazon AGI Lab \\
  \And
  Anne G. E. Collins \\
  University of California, Berkeley \\
}

\begin{document}

\maketitle

\begin{abstract}
    Whether in agentic workflows, social studies, or chat settings, large language models (LLMs) are increasingly being asked to replace humans in choosing which goals to pursue, rather than completing predefined tasks. However, the assumption that LLMs accurately reflect human preferences for goal setting remains largely untested. We assess the validity of LLMs as proxies for human goal selection in a controlled, self-directed learning task borrowed from cognitive science. Across five models (GPT-5, Gemini 2.5 Pro, Claude Sonnet 4.5, Qwen3 32B, and Centaur), we find substantial divergence from human behavior. While people gradually explore and learn to achieve goals with diversity across individuals, most models exploit a single identified solution or show surprisingly low performance, with distinct patterns across models and little variability across instances of the same model. Chain-of-thought reasoning and persona steering provide limited improvements, and our conclusions hold across experimental settings. While they await confirmation in applied settings, these findings highlight the uniqueness of human goal selection and caution against its replacement with current models.
\end{abstract}

\section{Introduction}

As modern artificial intelligence (AI) systems become more capable and easily accessible, people increasingly rely on them for various tasks \cite{tamkin2024clio, zhao2024wildchat, zheng2023lmsys}. Beyond automating simple tasks, AI agents are becoming partners in thought \cite{collins2024building}, engaging in activities that were thus far considered a uniquely human prerogative. 
Crucially, we are progressively resorting to AI not only to help us \textit{complete} tasks and reach predefined goals, but also to \textit{select} which tasks and goals to pursue in the first place. 
In doing so, we are using AI as a proxy for ``autotelicity'', i.e., the ability to autonomously define goals \cite{colas2022autotelic}, which is considered key to flexible learning \cite{molinaro2025reward} and, more broadly, intelligent behavior \cite{chu2024praise, molinaro2023goal}. 
In this context, using AI to reduce mental effort \cite{risko2016cognitive} rests on the implicit assumption that conversational AI can substitute human intervention in a variety of settings. 
This assumption, in turn, stems from a fundamental inference problem: because intelligent chatbots display human-like linguistic abilities, we often attribute them anthropomorphic features that they inherently lack \cite{peter2025benefits}.

To determine which attributions are misguided, recent years have seen a fervent rise in benchmarking efforts to directly compare the outputs of large language models (LLMs) with human behavior (e.g., \cite{lampinen2024language, hendrycks2020measuring}). 
These efforts, however, overwhelmingly focus on estimating LLM capabilities -- what they can do -- while critical safety concerns typically originate from their propensities -- what they do when granted full autonomy \cite{mazeika2025utility, romero2026capabilities, summerfield2025lessons}. 
In a notorious case, the National Eating Disorders Association had to suspend its wellness chatbot, ``Tessa'', after it began proactively suggesting weight-loss goals and activities to users suffering from eating disorders \cite{aratani2023us}.
While this scenario depicts an extreme negative consequence, letting out-of-the-box AI systems select tasks for us could have important implications in a variety of domains, including which career to pursue or who to marry, where people have a demonstrated tendency to take AI tools' advice \cite{luettgau2025people}.

The risks extend beyond individual users. 
Researchers and policy-makers are increasingly using AI agents to model people's behavior or directly replace survey responders, assuming that ``silicon subjects'' will exhibit human-like biases and choices \cite{bisbee2024synthetic, chu2023language, hewitt2024predicting, sucholutsky2025using}. If LLMs provide inaccurate models of human goal selection, this practice could result in false conclusions about human cognition, hence faulty applications such as misguided legislation. 
Moreover, LLMs are being integrated into artificial systems for scientific discovery and self-directed learning architectures, based on the idea that they can substitute human judgments of interestingness \cite{faldor2024omni, mishra2025matter, mitchener2025kosmos, lu2026towards}. While this practice has yielded impressive results, it fails to explicitly test the validity of LLMs as a proxy for human goal selection -- a key aspect of intrinsic motivation \cite{oudeyer2007intrinsic}. 

Here, we address this issue by testing whether LLMs exhibit human-like goal selection in a controlled environment with self-determined goals \cite{molinaro2024latent}. 
We find that different models exhibit distinct behavioral signatures, but none fully capture the richness or range of human exploration. 
We demonstrate that our results hold regardless of the specific semantics of the environment and are robust to simple manipulations. 
These findings caution against the growing trend of using LLMs as proxies for human measures of interestingness, whether in personal tool use or scientific and policy applications.

\section{Related Work}

\subsection{Language Models as Silicon Subjects}

Recent work demonstrates a growing interest in using LLMs as proxies for people in social science studies, with researchers investigating whether these ``silicon subjects'' can replicate patterns of human cognition and behavior. 
This line of work is motivated by two complementary aims. On the one hand, there is value in replacing human respondents with cheaper, more easily accessible language models, although the practice remains debated \cite{aher2023using, bisbee2024synthetic, dillion2023can, demszky2023using, hagendorff2023machine, harding2024ai, hendrycks2020measuring}. On the other hand, behavioral experiments from the study of animal cognition can serve as a method for understanding and benchmarking machine intelligence \cite{binz2023using, rahwan2019machine}.
Studies comparing LLM and human behavior in matching experimental setups have yielded mixed results, with models replicating some aspects of human cognition while diverging substantially on others (e.g., \cite{lampinen2024language, strachan2024testing}). 

A key challenge in simulating people's behavior with LLMs is not only to predict the average response of a population, but also to adequately capture its variability \cite{bisbee2024synthetic, sorensen2024roadmap}. Failing to replicate this distributional spread risks modeling a falsely homogenized, median participant population, erasing the polarized subgroups and minorities that frequently drive real-world social dynamics \cite{argyle2023out, santurkar2023whose}.
The CogBench suite \cite{coda2024cogbench} represents a systematic effort to evaluate LLMs across multiple cognitive domains, revealing substantial variation both across models and across task types. 
This work has prompted the development of specialized models, including Centaur \cite{binz2025foundation}, which claims to be a ``foundation model of human psychology'' and to predict human behavior better than ad hoc cognitive models (but see \cite{orr2025not, xie2025centaur}). 

However, most tasks LLMs have been tested on so far share a critical limitation, from which human studies also suffer (with few exceptions, e.g. \cite{holton2024goal, molinaro2024latent, poli2022contributions, ten2021humans}): they measure performance based on a goal imposed by the experimenter, rather than studying goal selection itself \cite{horowitz2025llm, li2025large, pan2022effects}. 
This gap is particularly important given the increasing use of LLMs to model humans in domains from public policy to scientific research. 
Understanding self-directed behavior in LLMs requires studying not just how they reach assigned goals, but also which goals they autonomously select \cite{molinaro2023goal}. 
Here, we address this by comparing humans and LLMs with intrinsically motivated goal selection as the primary dependent variable.

\subsection{LLMs as Goal Selectors}

While traditional AI systems are optimized for predefined metrics, self-directed learning systems must identify their own objectives \cite{oudeyer2007intrinsic, schmidhuber2010formal}.
In the machine learning literature, this challenge has been addressed through various forms of intrinsic motivation and curiosity-driven exploration. 
Early work proposed that the external feedback provided to agents by the environment could be augmented by auxiliary intrinsic rewards corresponding, e.g., to novelty and surprise \cite{pathak2017curiosity, burda2018large}. 
Later approaches proposed instilling goal generation mechanisms based on learning progress directly into ``autotelic'' agents that find it rewarding to approach self-proposed goals \cite{colas2019curious, forestier2022intrinsically}. 
A key challenge in these systems is identifying tasks that are both learnable and interesting. To circumvent the problem, some have proposed querying foundation models such as LLMs for new and interesting problems to solve. 
The idea behind this approach is that foundation models have internalized notions of interestingness from human data \cite{faldor2024omni, zhang2024omni}. 
However, this assumption has not been empirically validated through direct comparison of LLM and human goal selection. 
Concurrent work \cite{lu2026mind} took a step in this direction by contrasting human and GPT-4o task generation in a static elicitation setting, but a controlled, dynamic evaluation across multiple models -- capturing how goal selection unfolds over time and varies across individuals -- is still missing.
The implications of human-AI alignment in goal selection extend beyond toy settings. In personal domains, people readily follow the advice of chatbots, even when it does not prove helpful for their particular circumstances \cite{luettgau2025people}. The consequences of this tendency are likely to be exacerbated when applied to goal selection rather than task completion. Moreover, as foundation models become integrated into automated science frameworks \cite{lu2026towards, mitchener2025kosmos}, they could eventually steer the academic discourse and drive innovation in ways that no longer align with human objectives. Given such responsibilities, it is important to characterize the goal selection patterns of foundation models.
Here, we test LLM goal setting in a simple, controlled experiment where behavior can be thoroughly characterized and compared with human choices. 

\section{Methods}\label{sec:methods}
To study the goal selection tendencies of conversational AI, we evaluated the output of different LLMs in an iterative goal-contingent learning task and compared their choices to those of 175 human participants in an interactive version of the same environment \cite{molinaro2024latent}.
Our analysis code is available at \texttt{https://github.com/gaiamolinaro/llm-goal-selection}. Our data is available upon request.

\subsection{Task}

\begin{wrapfigure}[15]{r}{0.5\textwidth}
    \centering
    \vspace{-19pt}
    \includegraphics[width=0.5\textwidth]{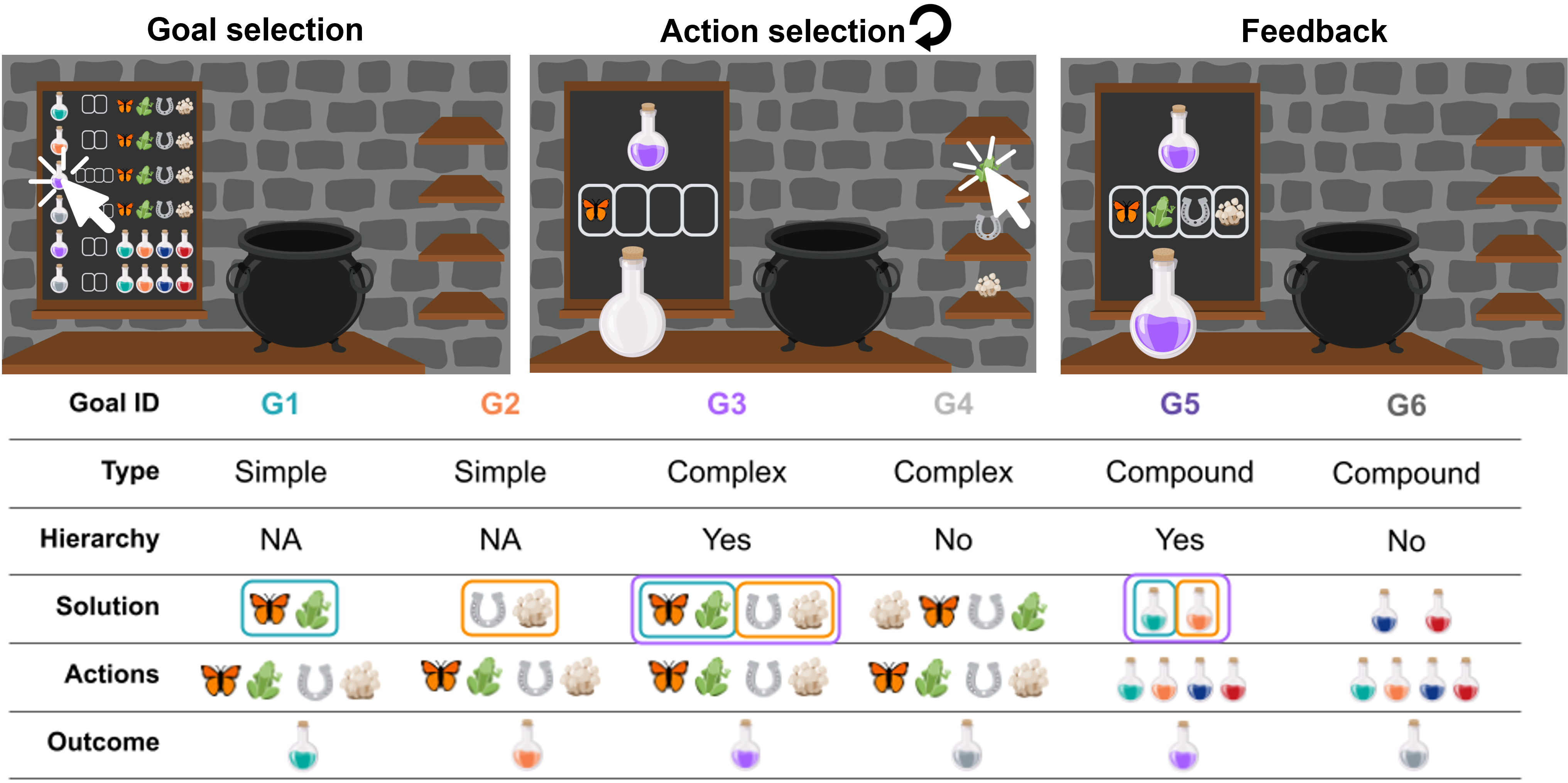}
    \caption{\textbf{Task structure}. Top: screenshots from the interactive version of the task developed for human participants. 
    Bottom: schematic representation of goals and their characteristics, with hierarchical relationships highlighted. Reproduced with permission from \cite{molinaro2024latent}.}
    \label{fig:task}
\end{wrapfigure}

Our LLM goal selection task was adapted from \cite{molinaro2024latent}. 
This environment enables the study of goals as the dependent variable of interest, rather than a setting defined by the experimenter and imposed on participants. 
By limiting the number of goals to six distinct options, this task addresses the open-ended question of goal selection while keeping quantitative analyses tractable and excluding potential confounds.
In the original experiment, human participants interacted with a computerized task presented as an ``alchemy game'', in which individuals took the role of ``apprentices'' (Figure \ref{fig:task}). 
Here, a goal is defined as brewing a specific potion, which is achieved by selecting a specific sequence of ingredients. 
On each trial, participants first selected their goal by indicating one of the available potions. 
Then, they were asked to pick a specific number of ingredients in order. 
Finally, they received deterministic feedback indicating whether the selected ingredients were added to the cauldron in the correct order, yielding the goal potion, or not (empty flask). The correct recipe for each potion was static and predefined, but initially unknown to participants. 
When selecting a potion, participants could leverage available information about the number (either two or four) and the type of ingredients required by it (either ``basic'' ingredients or other, pre-made potions). 
Two aspects of the goal space were manipulated: difficulty and hierarchical structure. 
The first factor reflected the number of ingredients required by a potion and was therefore known to participants. 
The second factor depended on the fact that a subset of the potions shared hidden common structures with others, such that identifying the correct recipe for the former could help solve the latter; however, some potions shared no hierarchical dependencies with others. 
The exact correspondence between potion identity, solution, color, and position on the screen, and the identity and order of ingredients, were randomized by creating 10 different task configurations used across participants as in \cite{molinaro2024latent}. 
To familiarize themselves with the task environment, participants completed a practice stage with forced goals (two iterations per goal in random order). 
Then, participants completed six blocks of 24 trials each, with free choices for both goal and action selection. 
Finally, they were presented with a surprise test, where each potion was presented four times, and participants had to make their best guess for the correct recipe. 
The test phase was necessary to measure individuals' acquired knowledge independent of goal selection. 
For instance, a participant who chose the same goal at which they succeeded for the entire duration of the task would show perfect performance during learning, but would fail at testing. 
In addition to the potions that were available during the main task (i.e., in-distribution), the test phase contained two out-of-distribution potions that could not be selected during the learning phase but whose correct recipe could be inferred from the other potions' solutions. 
Participants were not told about the final test ahead of time, nor were they given additional payment or course credits for learning any of the potions' recipes. 
Therefore, any observed efforts to learn were largely intrinsically motivated.
We refer the reader to \cite{molinaro2024latent} for details.

We turned the experiment into a procedurally generated text-based format suitable for LLM inference (Appendix \ref{app:prompt}). At each step, the model's choices were appended to the next step's prompt, such that the LLM could learn in context from its interaction history. 
Similar to the human participants' task, our LLM adaptation involved multi-turn interactions, where the model was prompted to first select a goal and then, contingent on its selection, a series of ingredients. 

While a modality mismatch between LLM and human data can be seen as a limiting factor, multiple reasons support this choice. First, prior work established text-based adaptations of interactive cognitive tasks as a valid method for probing LLM behavior and cognitive biases (e.g., \cite{binz2023using, coda2024cogbench}). Text is also the only available input modality for a subset of the models we evaluate, and interactive tasks for LLMs would introduce additional confounds due to limitations in their perceptual capabilities. Moreover, the text-based format reflects how LLMs typically receive instructions and operate in practice, which makes results more applicable to real-world deployment. Consequently, observing LLM behavior under favorable conditions would strengthen rather than undermine any conclusion about their divergence.
To verify that modality differences were not the sole driver of our results, we also present data from GPT-5.4, which -- through its computer-using capabilities -- was directly connected to an online platform matching the human experiment exactly. Our conclusions were not affected (Appendix \ref{app:interactive}).

\begin{figure}[t!]
    \centering
    \includegraphics[width=0.9\linewidth]{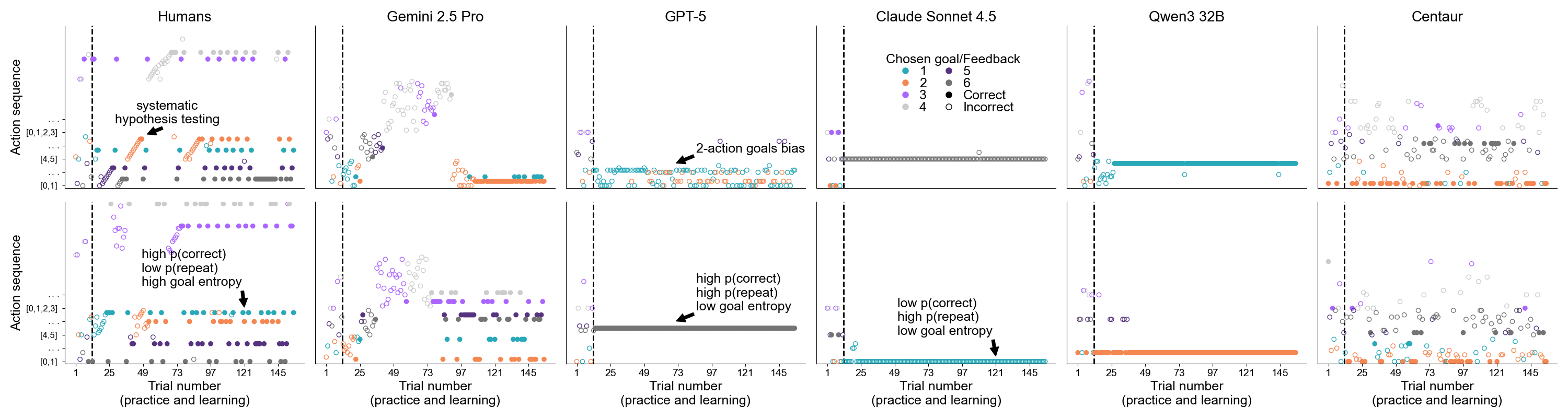}
    \caption{
      \textbf{Example action and goal selection choices}. 
      Each subplot represents a single participant/simulation (two examples per type to illustrate variability) over the practice and learning phases of the task (separated by a dotted line).
      Each dot shows the index of the particular sequence of actions selected. 
      Ingredients representing pre-made potions were labeled 4-7 for clearer visualization. 
    }
    \label{fig:example_actions}
\end{figure}

\begin{figure}[t!]
    \centering
    \includegraphics[width=0.9\linewidth]{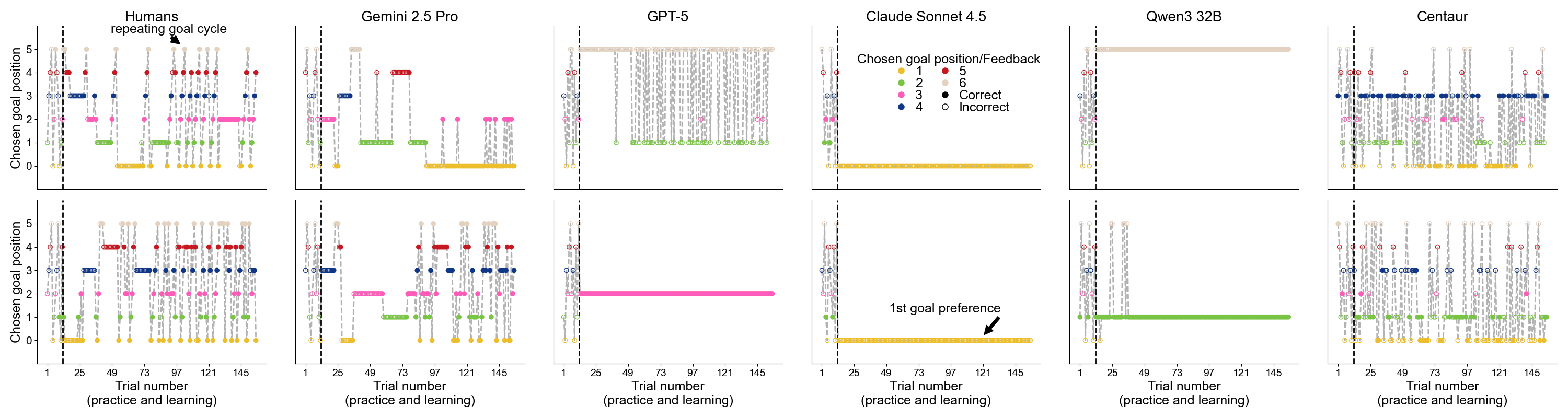}
    \caption{
      \textbf{Example goal position choices in humans and models}.
      Each column illustrates the goal selection of a single participant/simulation (indexed and color-coded) over trials.
    }
    \label{fig:example_goals}
\end{figure}

\subsection{Models}
We compared the output of various LLMs to the behavior of human participants, re-analyzed from \cite{molinaro2024latent}. The models covered a range of architectures, met our computational limitations, and were considered state-of-the-art at the time of data collection: GPT-5 (OpenAI), Gemini 2.5 Pro (Google), Claude Sonnet 4.5 (Anthropic), and Qwen3 32B (Qwen). We also collected responses from Centaur, a specialized model fine-tuned on psychology experiments \cite{binz2025foundation}.
We collected each model's responses in 50 separate iterations of the task, using each of the 10 task configurations approximately five times. 
For GPT-5, we used minimal reasoning effort and low verbosity settings; temperature was fixed to 1 by design. For all other models, we set a temperature of 1\footnote{For Centaur, we note that this setting often required multiple retries with the same prompt to get valid outputs (i.e., decisions that matched one of the available options).}, but present additional results using a temperature of 0 in Appendix \ref{app:temp0}. Wherever relevant, we set the top-P parameter to 1. 

\subsection{Enabling Reasoning}
Performance often improves when the model is allowed to ``reason'' before answering, i.e., break down complex problems by simulating thinking patterns in language before providing a final answer \cite{kojima2022large, krishnamurthy2024can, wei2022chain}. To assess the impact of reasoning on model behavior in our task, we prompted a subset of models that have reasoning functionality (Gemini 2.5 Pro, GPT-5, and Qwen3 32B) to think through each trial's choices step by step (\cite{coda2024cogbench}; Appendix \ref{app:prompt_cot}) to enable chain-of-thought (CoT). 
This procedure made the models elaborate about various options before giving a final answer. 

\subsection{Steering Models with Human Personas}
LLMs are known to be sensitive to the specific prompt they are instructed with \cite{sahoo2024systematic}. 
Wording that instructs the LLM to align with a persona or demographic category can result in responses that better align with the specified category of people \cite{bisbee2024synthetic, white2023prompt}.
In a follow-up experiment, we thus tested two powerful models, Gemini 2.5 Pro and GPT-5, on their ability to output more human-like responses following a prompt that explicitly stated the experimental conditions for the human subjects in the original study, including their university student status (Appendix \ref{app:prompt_persona}). 

\subsection{Manipulating Semantics and Experimental Paradigm}
Variants of the alchemy game presented here have a rich tradition in cultural evolution research, partly because its structure generalizes to real-world innovation and problem-solving \cite{derex2016partial, pan2025large, smaldino2024maintaining}. 
Our core results focus on this framing, which enables direct comparison to human data. 
To probe the generalizability of our results, we ran two follow-up experiments where the structure of the task is maintained (including its compositional features), but the identity of goals and actions is changed.
In the first variant (``skill acquisition''; Appendix \ref{app:prompt_skill_acquisition}), potions were replaced with ``acrobatic moves'' named after fictional athletes and ingredients with ``actions'' such as ``arch'' or ``tuck'' to reflect a slightly more naturalistic scenario.
In the second variant (``non-words''; Appendix  \ref{app:prompt_nonwords}), all key elements of the game were replaced by invented words from the developmental psychology literature (e.g., ``daxes''  and ``blickets''; \cite{gopnik2000detecting, markman1984children}). This abstracted the semantics away from the game, testing the generalizability of our findings to different domains.
The original task from \cite{molinaro2024latent} is also one of the very few human experiments where goals are the dependent variable of interest. To further validate our findings in alternative settings, we test the same models on a completely separate paradigm taken from \cite{holton2024goal} and compare their behavior to that of humans, as detailed in Appendix \ref{app:holton}. 

\subsection{Metrics and Data Analysis}

To characterize the diversity of human goal-directed learning, we compute the following metrics.
\paragraph{Performance.} 
Task choice accuracy -- the proportion of trials where the correct sequence of ingredients was selected, contingent on the current goal -- was calculated for different task phases.
  \begin{itemize}
    \item Overall learning performance: task choice accuracy across all learning trials.
    \item Blockwise learning performance: choice accuracy for each block of the learning phase.
    \item In-distribution test performance: choice accuracy in the test phase for the trials when a learning phase potion was externally set as the goal (four times each).
    \item Out-of-distribution test performance: the proportion of trials in which participants correctly inferred the correct recipes for two potions that were not available as targets in the learning phase, but for which the solution could be derived from knowledge about other potions.
  \end{itemize} 
Given the short duration of the practice phase, we focus on the learning and test phases in writing, but show practice phase performance in the plots for completeness. 

\paragraph{Goal selection.} We use several measures to characterize goal selection in the learning phase.
  \begin{itemize}
    \item Probability of choosing two-action goals: the proportion of trials in which a subject chose a two-action goal (easier), compared to a four-action goal (harder).
    \item Probability of repeating a goal: the proportion of trials in which a subject chose the same goal consecutively. 
    \item Goal selection entropy: the entropy of the empirical distribution of free goal choices (maximal when all goals are selected equally often), calculated using the natural logarithm over probabilities without applying normalization.
    \item Preferred goal position: the position (on the screen or in the text) of the most chosen goal. A score of 0 means that the participant's preferred potion was the first one (0-indexed) to appear on the screen (for humans) or in the written list of available goals (for LLMs).
    \item Goal cycles: the number of times a goal cycle (e.g., goals 0 through 5) was performed, taking the maximum repetition count over any permutation of the six goals.
  \end{itemize}
\paragraph{Position-based systematic hypothesis testing.} The number of times participants tested possible solutions for the same goal based on the ingredients' positions on the screen, following a systematic sorting of possible sequences. 
We chose this measure to quantify participants' tendency to strategically explore possible goal solutions, selecting combinations of ingredients that matched the order in which ingredients were presented. 
For instance, human participants often tested action sequences [0,1], [0, 2], [0,3], [1, 2], etc., in this order, where each number represents the top (0) to bottom (3) location of the ingredient on the screen.
This strategy allows participants to avoid remembering every incorrect action sequence they tested, and instead only keep in mind their search algorithm and the last ingredient sequence tried. 
While LLMs do not have the same capacity limitations as humans in this task, similar biases could emerge from training on human data in other domains. 

\paragraph{Statistical tests.} A fully aligned LLM system should produce responses that are similar to humans on average and in distribution, reproducing inter-individual variability. To assess the similarity of human and AI data distributions for the above-listed metrics, we performed the Kolmogorov-Smirnov test for continuous data, and the $\chi^2$ contingency test for differences in discrete data.
To compare the multi-dimensional distributions for learning performance over time (i.e., block number), we used the Energy distance statistic. 
Wherever relevant, we also report the Mann-Whitney U test statistic for differences in central tendency between two independent samples. 

\begin{wraptable}[11]{r}{0.5\textwidth}
\vspace{-13pt}

\centering
\renewcommand{\arraystretch}{1.1}
\setlength{\tabcolsep}{3.5pt}
\scriptsize
\caption{\textbf{No model fully matches human goal selection.} Significant (red) and non-significant (green) differences between humans and model scores' medians and distributions. 
}
\label{tab:summary}
\begin{NiceTabular}{c|c|c|c|c|c}[hvlines]
\CodeBefore
    \SplitCell{2}{2}{red!30}{red!30}
    \SplitCell{2}{3}{red!30}{red!30}
    \SplitCell{2}{4}{red!30}{red!30}
    \SplitCell{2}{5}{red!30}{red!30}
    \SplitCell{2}{6}{red!30}{red!30}
    \SplitCell{3}{2}{green!30}{red!30}
    \SplitCell{3}{3}{red!30}{red!30}
    \SplitCell{3}{4}{red!30}{red!30}
    \SplitCell{3}{5}{red!30}{red!30}
    \SplitCell{3}{6}{red!30}{red!30}
    \SplitCell{4}{2}{green!30}{green!30}
    \SplitCell{4}{3}{red!30}{red!30}
    \SplitCell{4}{4}{red!30}{red!30}
    \SplitCell{4}{5}{green!30}{green!30}
    \SplitCell{4}{6}{red!30}{green!30}
    \SplitCell{5}{2}{green!30}{green!30}
    \SplitCell{5}{3}{red!30}{red!30}
    \SplitCell{5}{4}{red!30}{red!30}
    \SplitCell{5}{5}{red!30}{red!30}
    \SplitCell{5}{6}{green!30}{red!30}
    \SplitCell{6}{2}{green!30}{green!30}
    \SplitCell{6}{3}{red!30}{red!30}
    \SplitCell{6}{4}{red!30}{red!30}
    \SplitCell{6}{5}{red!30}{red!30}
    \SplitCell{6}{6}{red!30}{red!30}
    \SplitCell{7}{2}{red!30}{red!30}
    \SplitCell{7}{3}{red!30}{red!30}
    \SplitCell{7}{4}{red!30}{red!30}
    \SplitCell{7}{5}{red!30}{red!30}
    \SplitCell{7}{6}{red!30}{red!30}
    \SplitCell{8}{2}{red!30}{red!30}
    \SplitCell{8}{3}{green!30}{red!30}
    \SplitCell{8}{4}{red!30}{red!30}
    \SplitCell{8}{5}{red!30}{red!30}
    \SplitCell{8}{6}{red!30}{red!30}
    \SplitCell{9}{2}{green!30}{green!30}
    \SplitCell{9}{3}{red!30}{red!30}
    \SplitCell{9}{4}{red!30}{red!30}
    \SplitCell{9}{5}{red!30}{red!30}
    \SplitCell{9}{6}{red!30}{red!30}
    \SplitCell{10}{2}{red!30}{red!30}
    \SplitCell{10}{3}{red!30}{red!30}
    \SplitCell{10}{4}{red!30}{red!30}
    \SplitCell{10}{5}{red!30}{red!30}
    \SplitCell{10}{6}{red!30}{red!30}
\Body
    \diagbox{\textbf{Md}}{\textbf{Distr.}} & Gemini & GPT & Claude & Qwen & Centaur \\
    P(corr.) learn.      &   &   &   &   &   \\
    P(corr.) in-distr.   &   &   &   &   &   \\
    P(corr.) out-distr.  &   &   &   &   &   \\
    P(2-act. goals)      &   &   &   &   &   \\
    P(repeat goal)       &   &   &   &   &   \\
    Goal entropy         &   &   &   &   &   \\
    Pref. goal pos.      &   &   &   &   &   \\
    N. goal cycles       &   &   &   &   &   \\
    N. hyp. test.        &   &   &   &   &   \\
\end{NiceTabular}
\end{wraptable}
To quantify the similarity between human and LLM goal selection across features (normalized by the human mean and SD), we calculated the minimum Euclidean distance between each human and every other participant, then compared it to the distance between each model simulation and the closest human.

\section{Results}

Example data from individual humans and LLMs reveal several differences (Figures \ref{fig:example_actions}-\ref{fig:example_goals}) which we systematically test below (Table \ref{tab:summary}). 
Note that the human data were analyzed in detail in \cite{molinaro2024latent}. Here, we summarize the main results to establish comparisons with LLMs.

\subsection{Performance} 
\begin{figure}[t!]
    \centering
    \includegraphics[width=0.9\linewidth]{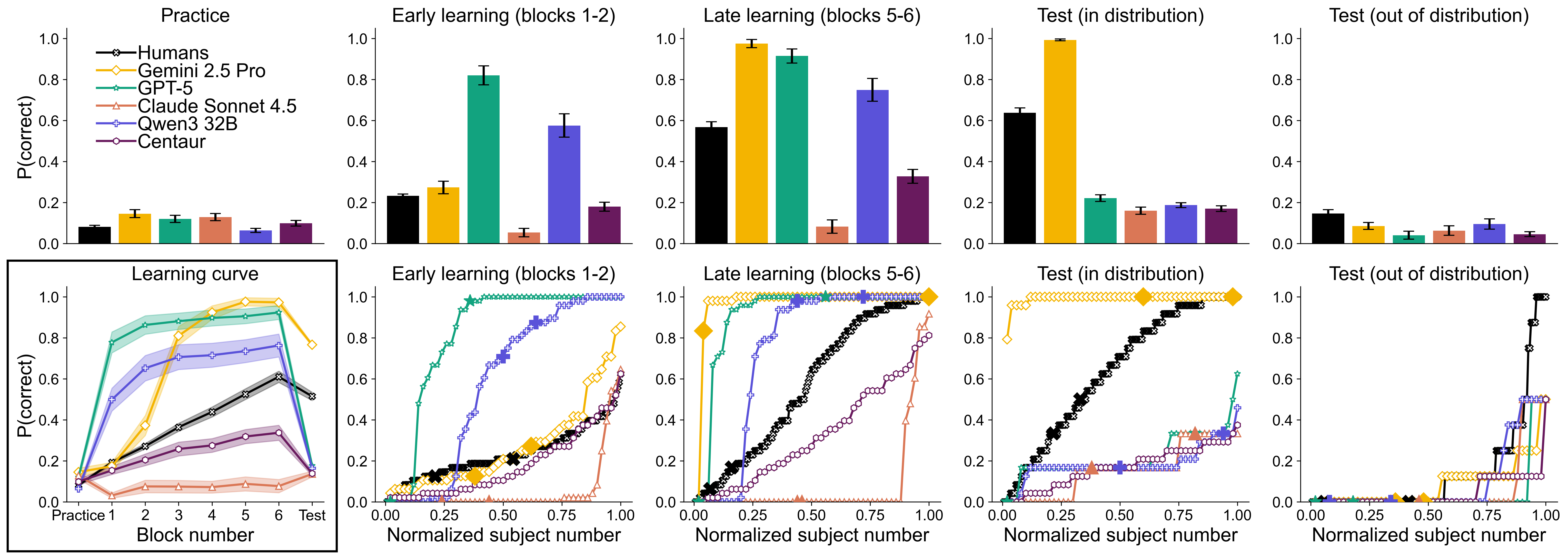}
    \caption{\textbf{Performance across task phases.} Top: average performance in the practice, early learning, late learning, and test blocks. Bottom, first subplot: learning curve. Bottom, following subplots: sorted individual participant scores, with the x-axis normalized by the number of participants, such that it represents the proportion of participants with a score equal to or lower than the current y. Error bars and shading indicate the S.E.M. Larger markers highlight examples from Figures \ref{fig:example_actions}-\ref{fig:example_goals}.}
    \label{fig:learning}
\end{figure}

Although highly variable, humans tended to progressively learn the solution for each goal, after which they appeared to rehearse each recipe in cycles. 
This resulted in relatively high performance during learning without drastic changes at testing, including some inference capabilities for out-of-distribution tasks (Figure \ref{fig:learning}). Models deviated from this pattern of results in a few ways.

On average, humans chose the correct combination of ingredients, contingent on their own goal selection, with a probability of 0.40 $\pm$ 0.22 (much higher than chance, i.e., 1/12 $\approx$ 0.08 for two-action goals, and 1/24 $\approx$ 0.04 for four-action goals).
Models' average performance was even higher than humans' for Gemini 2.5 Pro, GPT-5, and Qwen3 32B, but lower for Claude Sonnet 4.5 and Centaur (Tables \ref{tab:stats_descr_base_models}, \ref{tab:stats_human_vs_llm_base_models}). 
Both the overall distribution of performance scores and the blockwise learning progression of these models were different from those of human participants (all Energy distance measures $>$ 0.16, all p $<$ 0.001), failing to replicate the variability of the human data.

By contrast, most models' in-distribution test scores were \textit{lower} than humans' (M = 0.64 $\pm$ 0.32) and different in spread. 
Compared to humans (M = 0.15 $\pm$ 0.26), most models' out-of-distribution test scores were also lower and differently distributed (Tables \ref{tab:stats_descr_base_models}, \ref{tab:stats_human_vs_llm_base_models}). Centaur's distribution was not significantly different from humans' on out-of-distribution scores, but the median performance was significantly lower. 
Gemini 2.5 Pro was an exception, with near-perfect scores on the in-distribution test and near-human out-of-distribution test scores.  
GPT-5 and Qwen3 32B were particularly striking, having shown super-human performance at learning but significantly dropping at testing. 
These results are indicative of ``reward hacking'', i.e., exploiting known solutions to earn positive feedback -- even when not explicitly incentivized to do so \cite{amodei2016concrete}. 
By contrast, Claude Sonnet 4.5 showed surprisingly low performance throughout (Figure \ref{fig:learning}), a somewhat surprising result that is likely not attributable to parsing errors for two reasons. First, Claude reliably produced well-formed, valid responses, indicating intact instruction following. Second, Claude’s performance rose substantially in a semantic variant of the alchemy task (Appendix \ref{app:semantics}) despite an identical output format and task structure.

\subsection{Goal Selection}
In our setup, the key variables of interest relate to goal selection, as learning is self-directed and contingent on the free choice of a goal on each trial (Figure \ref{fig:goal_action_selection}). 
Compared to humans (M = 0.61 $\pm$ 0.14), GPT-5, Claude Sonnet 4.5, and Qwen3 32B were consistently biased towards simpler goals that required two ingredients compared to harder, four-ingredient potions. 
Gemini 2.5 Pro showed a similar level of preference for two-action goals as humans, both in average and distribution; Centaur, in average only (Tables \ref{tab:stats_descr_base_models}, \ref{tab:stats_human_vs_llm_base_models}).
Compared to humans (M = 0.54 $\pm$ 0.18), most models also showed a strong preference for re-selecting the same potion as in the previous trial -- though Centaur had the opposite tendency. 
Gemini 2.5 Pro once again stood out, resembling human average scores and distributions more closely.
However, like most other models, even Gemini 2.5 Pro's goal entropy was lower and differently distributed than humans' (M = 1.67 $\pm$ 0.14; Tables \ref{tab:stats_descr_base_models}, \ref{tab:stats_human_vs_llm_base_models}). 
Compared to humans, most models also showed a stronger tendency to choose the first potion listed on the options menu, indicating biases that were irrelevant to learning and not present in human; the exceptions were GPT-5, which showed no bias, and Qwen3 32B, which preferred later options (Tables \ref{tab:stats_descr_base_models}, \ref{tab:stats_human_vs_llm_base_models}).
By contrast, people showed a tendency to iterate over goals in consistent cycles (Figure \ref{fig:example_goals}, M = 1.98 $\pm$ 2.36), which we only detected in Gemini 2.5 Pro (Tables \ref{tab:stats_descr_base_models}, \ref{tab:stats_human_vs_llm_base_models}).
Together, these results suggest that humans and LLMs differ significantly on several aspects of goal selection. Gemini 2.5 Pro showed some notable exceptions, suggesting that some human-like goal selection might emerge with more sophisticated training.

\subsection{Action Selection}

\begin{figure}[t!]
    \centering
    \includegraphics[width=1\linewidth]{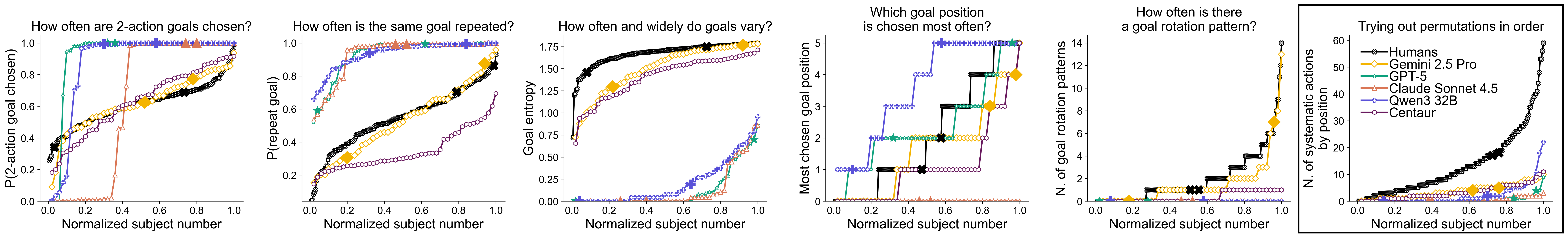}
    \caption{\textbf{Distributions of goal and action selection behaviors.} Sorted individual scores over the normalized subject number for various aspects of goal (first five subplots) and action selection (right-most subplot) in humans and models. Larger markers highlight examples from Figures \ref{fig:example_actions}-\ref{fig:example_goals}.}
    \label{fig:goal_action_selection}
\end{figure}

Many participants in the original experiment systematically tested various hypotheses for the correct recipe of each potion by following a pattern that matched the ingredients' order on the screen, e.g., testing the same two-action goal in consecutive trials with the first and second ingredient, first and third ingredient, first and fourth ingredient, and so on (ramping pattern in Figure \ref{fig:example_actions}).
Although other strategies were possible, this specific bias was the most prevalent and served as a target to compare to the models' goal selection patterns. 
On average, people followed such a strategic approach 13.36 $\pm$ 12.22 times. No model showed similar ordered exploration biases, with all distributions differing significantly from those of humans (Figure \ref{fig:goal_action_selection}; Tables \ref{tab:stats_descr_base_models}, \ref{tab:stats_human_vs_llm_base_models}).
This was the case even upon reducing the models' context window to approximate human memory limitations (Appendix \ref{app:cw}).

\subsection{Similarity to the Closest Human}
To quantify individual-level similarity between LLMs and humans across features, we performed a nearest-neighbors analysis. All models were significantly farther from the closest human than humans were from the closest neighbor in the feature space (all Mann-Whitney U $>$ 5656, all p $<$ 0.002; Figure \ref{fig:nearest_neighbor}), suggesting that LLMs did not match the behavior of any particular human in our dataset. 

\subsection{Impact of Reasoning} \label{sec:cot} 
In an additional experiment, models were encouraged to use chain-of-thought (CoT) reasoning. This manipulation had little impact on our results (Appendix \ref{app:prompt_cot}). However, we note a few significant differences. 
Learning performance (Figure \ref{fig:cot_learning}) increased relative to baseline, reaching super-human scores with tighter distributions in all models. Gemini 2.5 Pro, whose performance was already near-perfect on in-distribution tests at baseline, was even more consistent with CoT, suggesting it had explored, discovered, and recognized the solution for each goal in every iteration of the game. The drop in performance between learning and testing noted at baseline for GPT-5 and Qwen3 32B was somewhat reduced by CoT (Tables \ref{tab:stats_descr_cot_models}, \ref{tab:stats_human_vs_llm_cot_models}). GPT-5's out-of-distribution scores were similar to humans'.
CoT prompting also affected certain aspects of goal selection in Gemini 2.5 Pro (Figure \ref{fig:cot_goal_action_selection}). CoT reduced Gemini 2.5 Pro’s probability of repeating the same goal, which dropped below human levels (Tables \ref{tab:stats_descr_cot_models}, \ref{tab:stats_human_vs_llm_cot_models}).
Gemini 2.5 Pro also showed stronger tendencies to rotate through goals in consistent cycles with CoT, with an above-human average count (Tables \ref{tab:stats_descr_cot_models}, \ref{tab:stats_human_vs_llm_cot_models}).
One possible explanation for this pattern of results is that CoT helped the model explore and discover correct solutions faster, leaving more time for alternating through goals to practice before the end of the experiment. 
By contrast, goal selection patterns were largely consistent with our initial results for GPT-5. 
Qwen3 32B also remained consistent, but goal position biases faded away on average (Figure \ref{fig:cot_goal_action_selection}).

\subsection{Impact of Persona Steering} \label{sec:persona}
To steer models towards human signatures of behavior, we ran a separate version of the experiment with a prompt instructing models to act as human participants (Appendix \ref{app:prompt_persona}).
Such persona steering was not particularly effective in making models' responses more human-like in our setup, with minimal effects varying across models and measures (Figure \ref{fig:persona_learning}, Tables \ref{tab:stats_descr_base_models_persona}, \ref{tab:stats_human_vs_llm_base_models_persona}). 
One of the few notable differences was Gemini 2.5 Pro's goal selection, which became less repetitive than humans' while having a higher and more consistent incidence of cycling through goals. These results suggest that persona steering has variable effects based on the target model and metric.

\subsection{Impact of Semantics and Experimental Paradigm}
Our results held even in a separate ``skill acquisition'' setting, or after removing semantic priors from goals and actions by using ``non-words'' (Appendix \ref{app:prompt_skill_acquisition}-\ref{app:prompt_nonwords}; Figures \ref{fig:semantics_performance}-\ref{fig:semantics_goal_selection}).
The only meaningful difference was an increase in Claude 4.5 Sonnet's performance at learning and in-distribution testing in the skill acquisition, driven by improved hypothesis testing (Table \ref{tab:stats_descr_base_models_acrobat}). However, Claude's learning performance did not increase in the non-word scenario (Table \ref{tab:stats_descr_base_models_nonwords}).
Our conclusions are also not specific to the task structure presented in \cite{molinaro2024latent} or the above-listed metrics. By prompting models to complete an entirely different paradigm first introduced by \cite{holton2024goal}, we find that most models also differ from humans on separate goal-selection metrics, such as goal persistence (Appendix \ref{app:holton}). Gemini 2.5 Pro was a notable exception, showing similar biases to human participants. 

\section{Conclusion}

Most current benchmarks for LLMs test the ability to complete predefined tasks, but not model propensities with respect to goal selection itself. To begin addressing this gap, we borrowed an experimental paradigm from cognitive science and adapted it for LLM use. This allowed us to quantitatively compare human and model goal selection behavior in a controlled setting. By testing five state-of-the-art models across multiple behavioral dimensions, we sought to evaluate whether LLMs can serve as valid proxies for human intrinsic motivation. 

Overall, we find a strong disparity between humans' and LLMs' performance in the task. 
While people typically engaged in broad and progressive learning, some models showed reward hacking, while others had poor performance across all phases despite their general capabilities.
Gemini 2.5 Pro neared or even surpassed human performance, but failed to capture the diversity of scores seen in human participants. 
Differences between human and model behaviors also apply to goal selection. Compared to people, most models (except Gemini 2.5 Pro and Centaur) had a stronger bias to select easier goals and repeat the same goal selection in consecutive trials. 
Most models also showed a preference for the first available goal, which was not present in humans, and failed to replicate human tendencies to test possible solutions systematically. 
Even Centaur, explicitly trained to emulate humans in experimental settings, poorly captured people's goal selection.
These results were robust to standard interventions such as chain-of-thought reasoning and persona steering, and were true even with perfect input modality matching between the human and the model tasks (Appendix \ref{app:interactive}). Model behavior was minimally affected by the semantic context of the task. Our conclusions held true in an entirely separate paradigm, suggesting broad applicability despite the controlled experimental setting.

Different modes of divergence could pose different risks when integrated into hybrid human-AI systems. For example, delegating goal selection to inadequate models could lead to poor results; leaving goal setting to myopic models could lead to short-term learning gains at the expense of long-term retention; and goal selection dictated by super-human models could leave users behind -- exacerbating the risks of over-reliance \cite{klingbeil2024trust, ullrich2021development}.
To improve goal selection alignment between humans and AI, researchers could fine-tune LLMs on ad hoc data, such as in-depth interviews about individuals' goal selection strategies \cite{chu2023language, park2023generative}, identify and inject steering vectors directly into the models \cite{kim2025linear}, or resort to advanced training techniques \cite{wang2026learning, wu2026humanlm, zhou2026odyssim}. However, these methods are currently inaccessible to na\"ive users.
We also note that matching human goal selection is not always desirable. 
The danger lies not in the divergence between human and artificial goal selection itself, but in ignoring the extent of such divergence and the contexts in which it operates.

We acknowledge some limitations of our work. 
First, while we tested five popular LLMs, newer or other providers' models might exhibit different behavioral patterns, requiring continuous evaluation.  
Second, since we prioritized working with a controlled experimental setting, our findings await validation in complementary paradigms designed to study real-world settings \cite{lum2025bias}. At the same time, our setup is directly applicable to the increasingly widespread practice of simulating human participants in experimental psychology tasks for piloting purposes \cite{binz2025foundation}. 
Third, while we focused on behavioral patterns, mechanistic investigations of models' internal representations could provide additional insights into the phenomena we observed.
Finally, while we focused on humans and LLMs in isolation, research on joint goal selection with hybrid human-AI systems would also be highly valuable.
Taken together, our results establish a proof of concept and a meaningful lower bound: if models fail to align in a simplified, favorable setting, this warrants serious caution before assuming alignment in more complex ones. 


\begin{ack}
We thank Pierre-Yves Oudeyer and Cédric Colas for initial comments on the experimental approach, and Bryan Silverthorn and David Luan for enabling this project at the Amazon AGI Labs. A.G.E.C. and G.M. are also partly supported by NSF Grant 2336466 awarded to A.G.E.C. 
\end{ack}

\bibliographystyle{unsrtnat}
\bibliography{ref}

\newpage
\appendix
\onecolumn
\section{Descriptive Statistics}
\label{tab:stats_descr}
\begin{table}[h!]
\centering
\caption{\textbf{Descriptive statistics for the main experiment.} Mean values $\pm$ SD, with medians in parentheses.}
\label{tab:stats_descr_base_models}
\scriptsize
\begin{tabular}{lllllll}
\toprule
Model & Humans & Gemini 2.5 Pro & GPT-5 & Claude Sonnet 4.5 & Qwen3 32B & Centaur \\
Metric &  &  &  &  &  &  \\
\midrule
P(corr.) learn & 0.40 ± 0.22 (0.40) & 0.70 ± 0.16 (0.72) & 0.87 ± 0.27 (1.00) & 0.07 ± 0.19 (0.00) & 0.68 ± 0.38 (0.89) & 0.26 ± 0.19 (0.21) \\
P(corr.) in-distr. & 0.64 ± 0.32 (0.71) & 0.99 ± 0.03 (1.00) & 0.22 ± 0.11 (0.17) & 0.16 ± 0.12 (0.17) & 0.19 ± 0.08 (0.17) & 0.17 ± 0.10 (0.17) \\
P(corr.) out-distr. & 0.15 ± 0.26 (0.00) & 0.09 ± 0.12 (0.00) & 0.04 ± 0.14 (0.00) & 0.06 ± 0.16 (0.00) & 0.10 ± 0.18 (0.00) & 0.04 ± 0.09 (0.00) \\
P(2-act goal) & 0.61 ± 0.14 (0.63) & 0.62 ± 0.17 (0.62) & 0.93 ± 0.24 (1.00) & 0.62 ± 0.47 (1.00) & 0.88 ± 0.29 (1.00) & 0.63 ± 0.21 (0.66) \\
P(repeat goal) & 0.54 ± 0.18 (0.56) & 0.53 ± 0.21 (0.54) & 0.93 ± 0.12 (0.99) & 0.93 ± 0.13 (0.99) & 0.94 ± 0.08 (0.98) & 0.35 ± 0.12 (0.30) \\
Goal entropy & 1.67 ± 0.14 (1.71) & 1.51 ± 0.27 (1.61) & 0.12 ± 0.22 (0.00) & 0.12 ± 0.24 (0.00) & 0.20 ± 0.25 (0.07) & 1.41 ± 0.24 (1.51) \\
Pref. goal pos. & 2.11 ± 1.77 (2.00) & 1.66 ± 1.41 (2.00) & 2.42 ± 1.33 (2.00) & 0.00 ± 0.00 (0.00) & 3.62 ± 1.56 (4.00) & 1.16 ± 1.27 (1.00) \\
N. goal cycles & 1.98 ± 2.36 (1.00) & 1.56 ± 2.34 (1.00) & 0.00 ± 0.00 (0.00) & 0.00 ± 0.00 (0.00) & 0.00 ± 0.00 (0.00) & 0.34 ± 0.47 (0.00) \\
N. hyp. test. & 13.36 ± 12.22 (10.00) & 3.62 ± 2.64 (3.00) & 0.54 ± 1.50 (0.00) & 0.46 ± 0.83 (0.00) & 2.24 ± 4.36 (0.00) & 3.38 ± 2.64 (3.00) \\
\bottomrule
\end{tabular}
\end{table}

\begin{table}[h!]
\centering
\caption{Descriptive statistics for the main experiment with CoT prompting.}
\label{tab:stats_descr_cot_models}
\scriptsize
\begin{tabular}{llll}
\toprule
Model & Gemini 2.5 Pro CoT & GPT-5 CoT & Qwen3 32B CoT \\
Metric &  &  &  \\
\midrule
P(corr.) learn & 0.79 ± 0.09 (0.78) & 0.99 ± 0.02 (1.00) & 0.85 ± 0.11 (0.89) \\
P(corr.) in-distr. & 1.00 ± 0.00 (1.00) & 0.32 ± 0.16 (0.25) & 0.38 ± 0.15 (0.38) \\
P(corr.) out-distr. & 0.07 ± 0.08 (0.00) & 0.05 ± 0.08 (0.00) & 0.03 ± 0.10 (0.00) \\
P(2-act goal) & 0.58 ± 0.10 (0.59) & 0.88 ± 0.32 (1.00) & 0.85 ± 0.30 (0.99) \\
P(repeat goal) & 0.30 ± 0.14 (0.29) & 0.98 ± 0.02 (0.99) & 0.69 ± 0.16 (0.70) \\
Goal entropy & 1.70 ± 0.09 (1.72) & 0.04 ± 0.07 (0.00) & 0.73 ± 0.36 (0.74) \\
Pref. goal pos. & 0.88 ± 1.56 (0.00) & 2.06 ± 1.45 (2.00) & 1.74 ± 1.23 (2.00) \\
N. goal cycles & 4.38 ± 4.05 (3.00) & 0.00 ± 0.00 (0.00) & 0.00 ± 0.00 (0.00) \\
N. hyp. test. & 1.16 ± 1.27 (1.00) & 0.02 ± 0.14 (0.00) & 0.86 ± 1.71 (0.00) \\
\bottomrule
\end{tabular}
\end{table}

\begin{table}[h!]
\centering
\caption{Descriptive statistics for the main experiment with persona steering.}
\label{tab:stats_descr_base_models_persona}
\scriptsize
\begin{tabular}{llll}
\toprule
Model & Gemini 2.5 Pro & GPT-5 & Qwen3 32B \\
Metric &  &  &  \\
\midrule
P(corr.) learn & 0.77 ± 0.21 (0.81) & 0.88 ± 0.28 (1.00) & 0.68 ± 0.36 (0.82) \\
P(corr.) in-distr. & 0.98 ± 0.10 (1.00) & 0.23 ± 0.13 (0.17) & 0.20 ± 0.09 (0.17) \\
P(corr.) out-distr. & 0.11 ± 0.21 (0.00) & 0.00 ± 0.02 (0.00) & 0.06 ± 0.15 (0.00) \\
P(2-act goal) & 0.55 ± 0.18 (0.61) & 0.92 ± 0.27 (1.00) & 0.87 ± 0.31 (1.00) \\
P(repeat goal) & 0.33 ± 0.32 (0.17) & 0.95 ± 0.11 (0.99) & 0.94 ± 0.08 (0.98) \\
Goal entropy & 1.56 ± 0.40 (1.76) & 0.08 ± 0.19 (0.00) & 0.20 ± 0.27 (0.04) \\
Pref. goal pos. & 1.84 ± 1.54 (1.50) & 2.46 ± 1.22 (2.00) & 2.82 ± 1.69 (3.00) \\
N. goal cycles & 10.28 ± 7.47 (12.00) & 0.00 ± 0.00 (0.00) & 0.00 ± 0.00 (0.00) \\
N. hyp. test. & 1.38 ± 2.47 (0.00) & 0.72 ± 2.43 (0.00) & 1.34 ± 2.48 (0.00) \\
\bottomrule
\end{tabular}
\end{table}

\begin{table}[h!]
\centering
\caption{Descriptive statistics for the main experiment with the temperature parameter set to 0.}
\label{tab:stats_descr_base_models_temp0}
\scriptsize
\begin{tabular}{lllll}
\toprule
Model & Gemini 2.5 Pro & Claude Sonnet 4.5 & Qwen3 32B & Centaur \\
Metric &  &  &  &  \\
\midrule
P(corr.) learn & 0.58 ± 0.33 (0.76) & 0.06 ± 0.20 (0.00) & 0.48 ± 0.44 (0.62) & 0.56 ± 0.50 (1.00) \\
P(corr.) in-distr. & 0.95 ± 0.09 (1.00) & 0.15 ± 0.12 (0.17) & 0.13 ± 0.09 (0.17) & 0.14 ± 0.14 (0.17) \\
P(corr.) out-distr. & 0.15 ± 0.28 (0.00) & 0.05 ± 0.15 (0.00) & 0.13 ± 0.22 (0.00) & 0.00 ± 0.00 (0.00) \\
P(2-act goal) & 0.48 ± 0.25 (0.57) & 0.60 ± 0.49 (1.00) & 0.81 ± 0.36 (1.00) & 0.42 ± 0.49 (0.01) \\
P(repeat goal) & 0.61 ± 0.34 (0.81) & 0.97 ± 0.09 (0.99) & 0.97 ± 0.04 (0.99) & 0.99 ± 0.01 (0.99) \\
Goal entropy & 1.30 ± 0.45 (1.31) & 0.05 ± 0.15 (0.00) & 0.17 ± 0.25 (0.00) & 0.01 ± 0.02 (0.00) \\
Pref. goal pos. & 1.58 ± 1.61 (1.00) & 0.00 ± 0.00 (0.00) & 3.50 ± 1.87 (5.00) & 1.12 ± 1.63 (1.00) \\
N. goal cycles & 4.84 ± 6.13 (1.00) & 0.00 ± 0.00 (0.00) & 0.00 ± 0.00 (0.00) & 0.00 ± 0.00 (0.00) \\
N. hyp. test. & 6.00 ± 6.61 (3.50) & 0.16 ± 0.54 (0.00) & 1.42 ± 2.29 (0.00) & 2.66 ± 4.24 (0.00) \\
\bottomrule
\end{tabular}
\end{table}

\begin{table}[h!]
\centering
\caption{Descriptive statistics for the skill acquisition experiment.}
\label{tab:stats_descr_base_models_acrobat}
\scriptsize
\begin{tabular}{llllll}
\toprule
Model & Gemini 2.5 Pro & GPT-5 & Claude Sonnet 4.5 & Qwen3 32B & Centaur \\
Metric &  &  &  &  &  \\
\midrule
P(corr.) learn & 0.59 ± 0.26 (0.67) & 0.95 ± 0.16 (1.00) & 0.45 ± 0.41 (0.31) & 0.80 ± 0.29 (0.95) & 0.31 ± 0.20 (0.31) \\
P(corr.) in-distr. & 0.96 ± 0.11 (1.00) & 0.25 ± 0.13 (0.19) & 0.36 ± 0.24 (0.33) & 0.23 ± 0.11 (0.19) & 0.14 ± 0.09 (0.12) \\
P(corr.) out-distr. & 0.09 ± 0.20 (0.00) & 0.05 ± 0.15 (0.00) & 0.04 ± 0.13 (0.00) & 0.01 ± 0.09 (0.00) & 0.08 ± 0.10 (0.00) \\
P(2-act goal) & 0.56 ± 0.22 (0.57) & 0.86 ± 0.35 (1.00) & 0.67 ± 0.46 (0.99) & 0.95 ± 0.18 (1.00) & 0.72 ± 0.17 (0.76) \\
P(repeat goal) & 0.60 ± 0.23 (0.62) & 0.99 ± 0.01 (0.99) & 0.96 ± 0.04 (0.98) & 0.94 ± 0.08 (0.97) & 0.39 ± 0.14 (0.37) \\
Goal entropy & 1.02 ± 0.47 (1.04) & 0.00 ± 0.01 (0.00) & 0.07 ± 0.08 (0.06) & 0.13 ± 0.15 (0.06) & 1.19 ± 0.27 (1.19) \\
Pref. goal pos. & 2.14 ± 1.65 (2.00) & 2.80 ± 1.52 (3.00) & 0.18 ± 0.55 (0.00) & 3.20 ± 1.62 (3.50) & 1.30 ± 1.49 (1.00) \\
N. goal cycles & 1.24 ± 1.82 (1.00) & 0.00 ± 0.00 (0.00) & 0.00 ± 0.00 (0.00) & 0.00 ± 0.00 (0.00) & 0.24 ± 0.43 (0.00) \\
N. hyp. test. & 5.02 ± 2.66 (5.00) & 0.66 ± 1.19 (0.00) & 7.34 ± 8.67 (3.50) & 1.06 ± 1.48 (1.00) & 9.72 ± 4.14 (10.00) \\
\bottomrule
\end{tabular}
\end{table}

\begin{table}[h!]
\centering
\caption{Descriptive statistics for the non-words experiment.}
\label{tab:stats_descr_base_models_nonwords}
\scriptsize
\begin{tabular}{llllll}
\toprule
Model & Gemini 2.5 Pro & GPT-5 & Claude Sonnet 4.5 & Qwen3 32B & Centaur \\
Metric &  &  &  &  &  \\
\midrule
P(corr.) learn & 0.64 ± 0.24 (0.73) & 0.95 ± 0.19 (1.00) & 0.17 ± 0.32 (0.00) & 0.72 ± 0.30 (0.85) & 0.37 ± 0.20 (0.35) \\
P(corr.) in-distr. & 0.98 ± 0.05 (1.00) & 0.28 ± 0.14 (0.27) & 0.32 ± 0.21 (0.33) & 0.22 ± 0.09 (0.17) & 0.20 ± 0.09 (0.21) \\
P(corr.) out-distr. & 0.12 ± 0.19 (0.00) & 0.02 ± 0.10 (0.00) & 0.10 ± 0.22 (0.00) & 0.08 ± 0.16 (0.00) & 0.06 ± 0.10 (0.00) \\
P(2-act goal) & 0.53 ± 0.22 (0.61) & 0.92 ± 0.27 (1.00) & 0.62 ± 0.46 (0.97) & 0.98 ± 0.07 (1.00) & 0.69 ± 0.19 (0.75) \\
P(repeat goal) & 0.49 ± 0.25 (0.45) & 0.99 ± 0.03 (0.99) & 0.94 ± 0.10 (0.99) & 0.87 ± 0.13 (0.91) & 0.41 ± 0.15 (0.37) \\
Goal entropy & 1.45 ± 0.39 (1.61) & 0.03 ± 0.09 (0.00) & 0.16 ± 0.30 (0.00) & 0.36 ± 0.34 (0.31) & 1.34 ± 0.31 (1.44) \\
Pref. goal pos. & 2.56 ± 1.72 (2.00) & 2.50 ± 1.46 (2.00) & 0.00 ± 0.00 (0.00) & 3.16 ± 1.54 (3.00) & 1.54 ± 1.40 (1.00) \\
N. goal cycles & 2.80 ± 3.69 (1.00) & 0.00 ± 0.00 (0.00) & 0.00 ± 0.00 (0.00) & 0.00 ± 0.00 (0.00) & 0.28 ± 0.45 (0.00) \\
N. hyp. test. & 5.02 ± 4.49 (4.00) & 0.44 ± 1.87 (0.00) & 4.62 ± 4.45 (3.00) & 2.04 ± 3.54 (1.00) & 3.32 ± 2.76 (3.00) \\
\bottomrule
\end{tabular}
\end{table}

\clearpage
\section{Human-LLM Comparison Statistics} 
\begin{table}[h!]
\centering
\caption{Human-AI comparison statistics for the main experiment. Md. = median. Dist. = distribution. Median statistics are for the Mann-Whitney U test. Distribution tests are for the Kolmogorov-Smirnov test for continuous data and the $\chi^2$ contingency test for categorical data. * p < 0.05, ** p < 0.01, *** p < 0.001}
\label{tab:stats_human_vs_llm_base_models}
\scriptsize
\begin{tabular}{lllllllllll}
\toprule
 & \multicolumn{2}{c}{Gemini 2.5 Pro} & \multicolumn{2}{c}{GPT-5} & \multicolumn{2}{c}{Claude Sonnet 4.5} & \multicolumn{2}{c}{Qwen3 32B} & \multicolumn{2}{c}{Centaur} \\
 \cmidrule(lr){2-3} \cmidrule(lr){4-5} \cmidrule(lr){6-7} \cmidrule(lr){8-9} \cmidrule(lr){10-11}
 & Md. & Dist. & Md. & Dist. & Md. & Dist. & Md. & Dist. & Md. & Dist. \\
Metric &  &  &  &  &  &  &  &  &  &  \\
\midrule
P(corr.) learn & 7581.0*** & 0.629*** & 7947.0*** & 0.831*** & 821.5*** & 0.863*** & 6471.0*** & 0.589*** & 2729.0*** & 0.331*** \\
P(corr.) in-distr. & 7923.0*** & 0.746*** & 1272.5*** & 0.706*** & 943.0*** & 0.766*** & 1078.0*** & 0.746*** & 1013.5*** & 0.766*** \\
P(corr.) out-distr. & 4261.0 & 0.103 & 2915.0*** & 0.354*** & 3203.5*** & 0.294** & 3748.0 & 0.174 & 3544.0* & 0.197 \\
P(2-act goal) & 4689.0 & 0.163 & 8160.0*** & 0.909*** & 5275.5* & 0.580*** & 7652.5*** & 0.834*** & 4888.0 & 0.286** \\
P(repeat goal) & 4153.5 & 0.134 & 8353.5*** & 0.831*** & 8312.0*** & 0.820*** & 8573.5*** & 0.860*** & 1623.0*** & 0.554*** \\
Goal entropy & 2824.0*** & 0.323*** & 1.0*** & 0.994*** & 2.0*** & 0.994*** & 3.0*** & 0.994*** & 1042.0*** & 0.651*** \\
Pref. goal pos. & 3813.0 & 35.67*** & 5010.0 & 38.89*** & 1025.0*** & 94.66*** & 6456.5*** & 33.21*** & 3084.5** & 14.53* \\
N. goal cycles & 3805.0 & 0.140 & 1175.0*** & 0.731*** & 1175.0*** & 0.731*** & 1175.0*** & 0.731*** & 2059.0*** & 0.406*** \\
N. hyp. test. & 1798.5*** & 0.489*** & 399.5*** & 0.826*** & 335.0*** & 0.866*** & 1061.0*** & 0.666*** & 1641.5*** & 0.517*** \\
\bottomrule
\end{tabular}
\end{table}

\begin{table}[h!]
\centering
\caption{Human-AI comparison statistics for the main experiment with CoT prompting.}
\label{tab:stats_human_vs_llm_cot_models}
\scriptsize
\begin{tabular}{lllllll}
\toprule
 & \multicolumn{2}{c}{Gemini 2.5 Pro} & \multicolumn{2}{c}{GPT-5} & \multicolumn{2}{c}{Qwen3 32B} \\
 \cmidrule(lr){2-3} \cmidrule(lr){4-5} \cmidrule(lr){6-7}
 & Md. & Dist. & Md. & Dist. & Md. & Dist. \\
Metric &  &  &  &  &  &  \\
\midrule
P(corr.) learn & 8293.0*** & 0.806*** & 8750.0*** & 1.000*** & 8460.5*** & 0.811*** \\
P(corr.) in-distr. & 8075.0*** & 0.846*** & 1864.5*** & 0.526*** & 2221.0*** & 0.537*** \\
P(corr.) out-distr. & 4246.5 & 0.143 & 3784.5 & 0.157 & 3057.5*** & 0.294** \\
P(2-act goal) & 3762.5 & 0.154 & 7700.0*** & 0.880*** & 7344.0*** & 0.789*** \\
P(repeat goal) & 1371.0*** & 0.569*** & 8750.0*** & 1.000*** & 6176.0*** & 0.326*** \\
Goal entropy & 4959.0 & 0.143 & 0.0*** & 1.000*** & 86.0*** & 0.929*** \\
Pref. goal pos. & 2427.5*** & 35.86*** & 4449.5 & 24.00*** & 4031.0 & 34.92*** \\
N. goal cycles & 6251.5*** & 0.294** & 1175.0*** & 0.731*** & 1175.0*** & 0.731*** \\
N. hyp. test. & 608.5*** & 0.806*** & 182.0*** & 0.940*** & 530.0*** & 0.786*** \\
\bottomrule
\end{tabular}
\end{table}

\begin{table}[h!]
\centering
\caption{Human-AI comparison statistics for the main experiment with persona steering.}
\label{tab:stats_human_vs_llm_base_models_persona}
\scriptsize
\begin{tabular}{lllllll}
\toprule
 & \multicolumn{2}{c}{Gemini 2.5 Pro} & \multicolumn{2}{c}{GPT-5} & \multicolumn{2}{c}{Qwen3 32B} \\
 \cmidrule(lr){2-3} \cmidrule(lr){4-5} \cmidrule(lr){6-7}
 & Md. & Dist. & Md. & Dist. & Md. & Dist. \\
Metric &  &  &  &  &  &  \\
\midrule
P(corr.) learn & 7844.0*** & 0.709*** & 7916.0*** & 0.860*** & 6638.5*** & 0.563*** \\
P(corr.) in-distr. & 7792.0*** & 0.766*** & 1304.5*** & 0.666*** & 1158.5*** & 0.746*** \\
P(corr.) out-distr. & 3927.0 & 0.114 & 2543.5*** & 0.414*** & 3256.0** & 0.274** \\
P(2-act goal) & 3617.0 & 0.280** & 8050.0*** & 0.920*** & 7396.0*** & 0.834*** \\
P(repeat goal) & 2355.5*** & 0.623*** & 8436.5*** & 0.883*** & 8595.0*** & 0.880*** \\
Goal entropy & 5513.0** & 0.383*** & 0.0*** & 1.000*** & 5.0*** & 0.994*** \\
Pref. goal pos. & 4052.5 & 4.88 & 5123.0 & 60.22*** & 5436.0** & 17.11** \\
N. goal cycles & 6790.5*** & 0.594*** & 1175.0*** & 0.731*** & 1175.0*** & 0.731*** \\
N. hyp. test. & 732.0*** & 0.720*** & 454.5*** & 0.806*** & 736.0*** & 0.706*** \\
\bottomrule
\end{tabular}
\end{table}

\begin{table}[h!]
\centering
\caption{Human-AI comparison statistics for the main experiment with the temperature parameter set to 0.}
\label{tab:stats_human_vs_llm_base_models_temp0}
\scriptsize
\begin{tabular}{lllllllll}
\toprule
 & \multicolumn{2}{c}{Gemini 2.5 Pro} & \multicolumn{2}{c}{Claude Sonnet 4.5} & \multicolumn{2}{c}{Qwen3 32B} & \multicolumn{2}{c}{Centaur} \\
 \cmidrule(lr){2-3} \cmidrule(lr){4-5} \cmidrule(lr){6-7} \cmidrule(lr){8-9}
 & Md. & Dist. & Md. & Dist. & Md. & Dist. & Md. & Dist. \\
Metric &  &  &  &  &  &  &  &  \\
\midrule
P(corr.) learn & 6076.0*** & 0.471*** & 712.0*** & 0.903*** & 4646.5 & 0.417*** & 4933.0 & 0.560*** \\
P(corr.) in-distr. & 7386.5*** & 0.606*** & 871.0*** & 0.766*** & 793.5*** & 0.794*** & 846.0*** & 0.766*** \\
P(corr.) out-distr. & 4029.5 & 0.114 & 3025.0*** & 0.334*** & 3993.5 & 0.154 & 2475.0*** & 0.434*** \\
P(2-act goal) & 3176.0** & 0.280** & 5250.0* & 0.600*** & 6994.5*** & 0.754*** & 3675.0 & 0.580*** \\
P(repeat goal) & 5154.0 & 0.463*** & 8579.0*** & 0.920*** & 8742.5*** & 0.989*** & 8750.0*** & 1.000*** \\
Goal entropy & 2698.0*** & 0.491*** & 0.0*** & 1.000*** & 1.0*** & 0.994*** & 0.0*** & 1.000*** \\
Pref. goal pos. & 3611.5 & 5.76 & 1025.0*** & 94.66*** & 6245.5*** & 42.52*** & 2922.0*** & 27.68*** \\
N. goal cycles & 4605.5 & 0.340*** & 1175.0*** & 0.731*** & 1175.0*** & 0.731*** & 1175.0*** & 0.731*** \\
N. hyp. test. & 2480.0*** & 0.317*** & 229.0*** & 0.886*** & 805.0*** & 0.646*** & 1306.5*** & 0.620*** \\
\bottomrule
\end{tabular}
\end{table}
\clearpage

\section{Nearest Neighbor Analysis}
\begin{figure}[h!]
    \centering
    \includegraphics[width=0.6\linewidth]{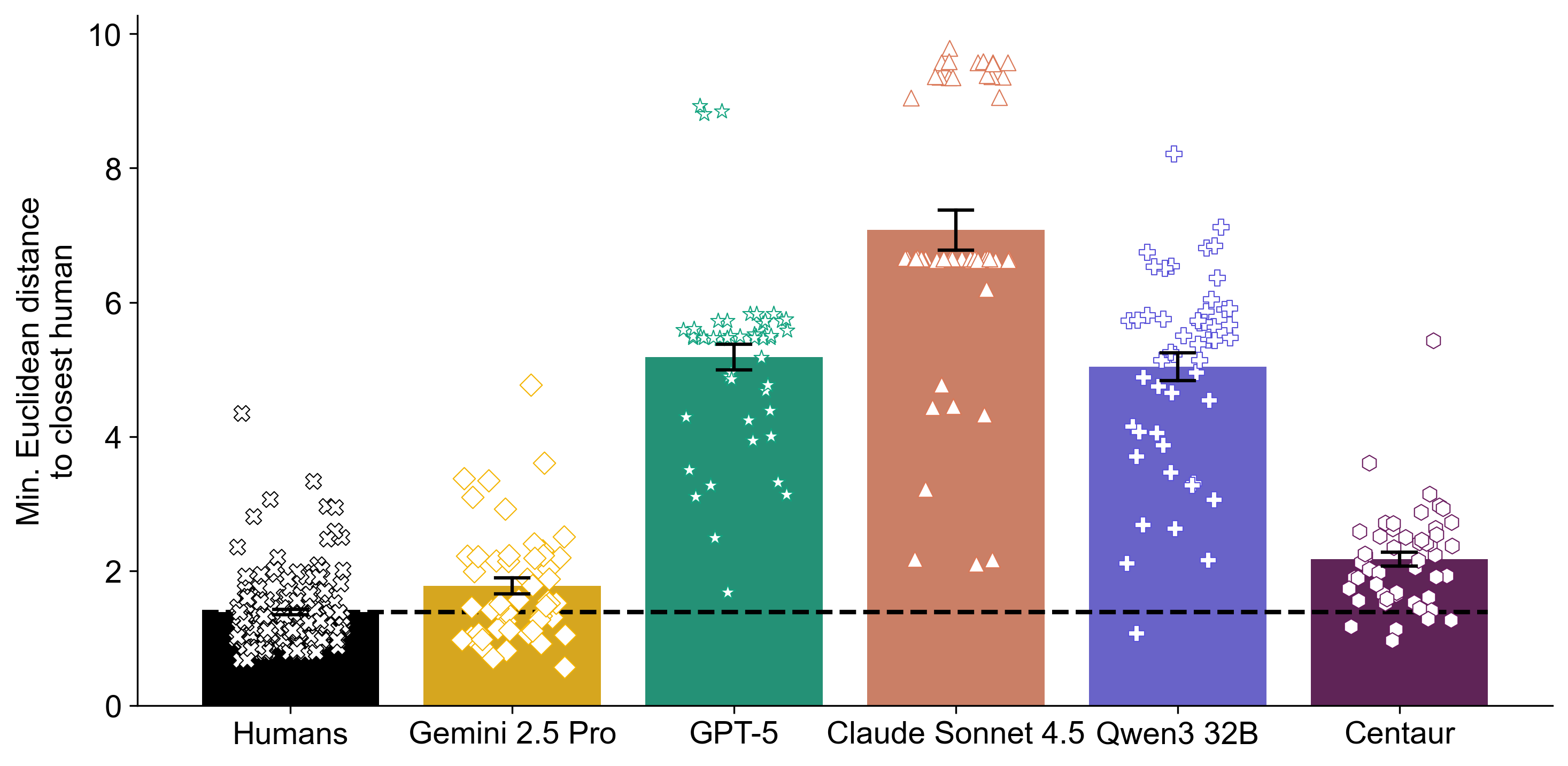}
    \caption{\textbf{LLMs differ from the closest human analog more than individual humans differ from each other.} Dots represent individual participants' or simulations' minimum Euclidean distance to any (other) human. Bars and error bars represent the mean and S.E.M.}
    \label{fig:nearest_neighbor}
\end{figure}

\newpage

\section{Effects of Reasoning} \label{app:cot}

To test the effects of reasoning on goal selection, a subset of the models was prompted to use a ``chain of thought'' before providing a final answer (Appendix \ref{app:prompt_cot}). 
A summary of the results is provided in Section \ref{sec:cot}.

\begin{figure}[h!]
    \centering
    \includegraphics[width=0.9\linewidth]{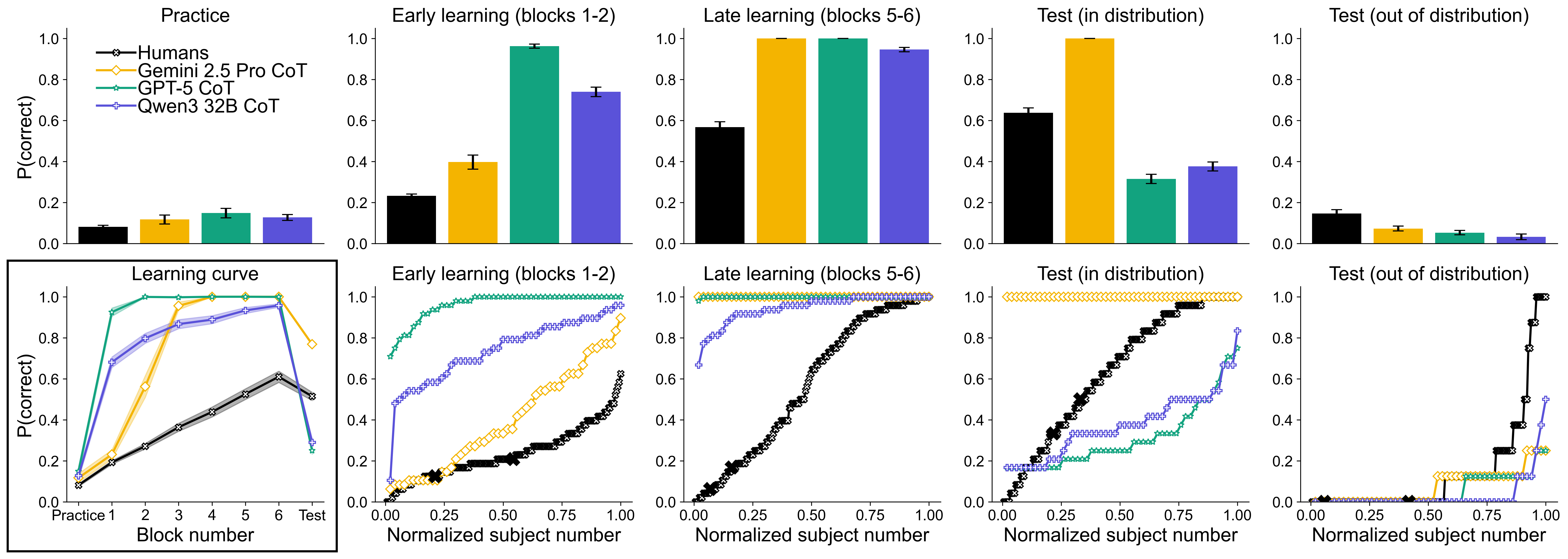}
    \caption{Performance across task phases for humans and models prompted with chain-of-thought inputs.}
    \label{fig:cot_learning}
\end{figure}

\begin{figure}[h!]
    \centering
    \includegraphics[width=1\linewidth]{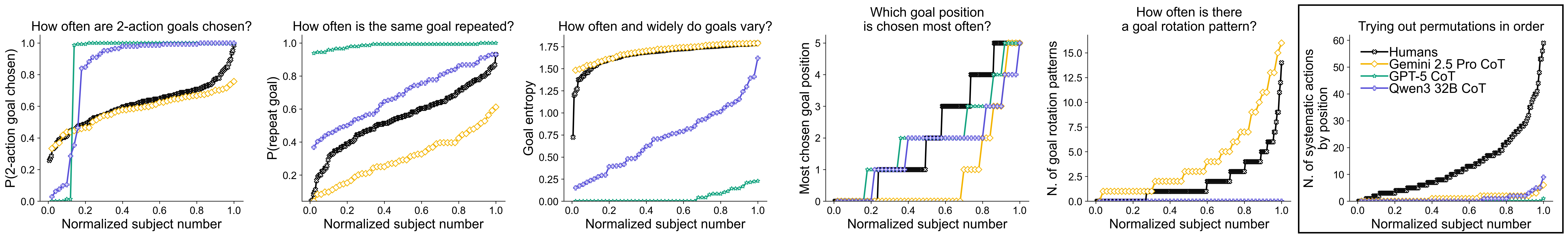}
    \caption{Distributions of goal and action selection behaviors in humans and models prompted with chain-of-thought inputs.}
    \label{fig:cot_goal_action_selection}
\end{figure}

\newpage

\section{Effects of Persona Steering} \label{app:persona}

To test the effects of steering through the description of a persona, a subset of the models was prompted to act like the original study's participants throughout the task (Appendix \ref{app:prompt_persona}). 
A summary of the results is provided in Section \ref{sec:persona}.

\begin{figure}[h!]
    \centering
    \includegraphics[width=0.9\linewidth]{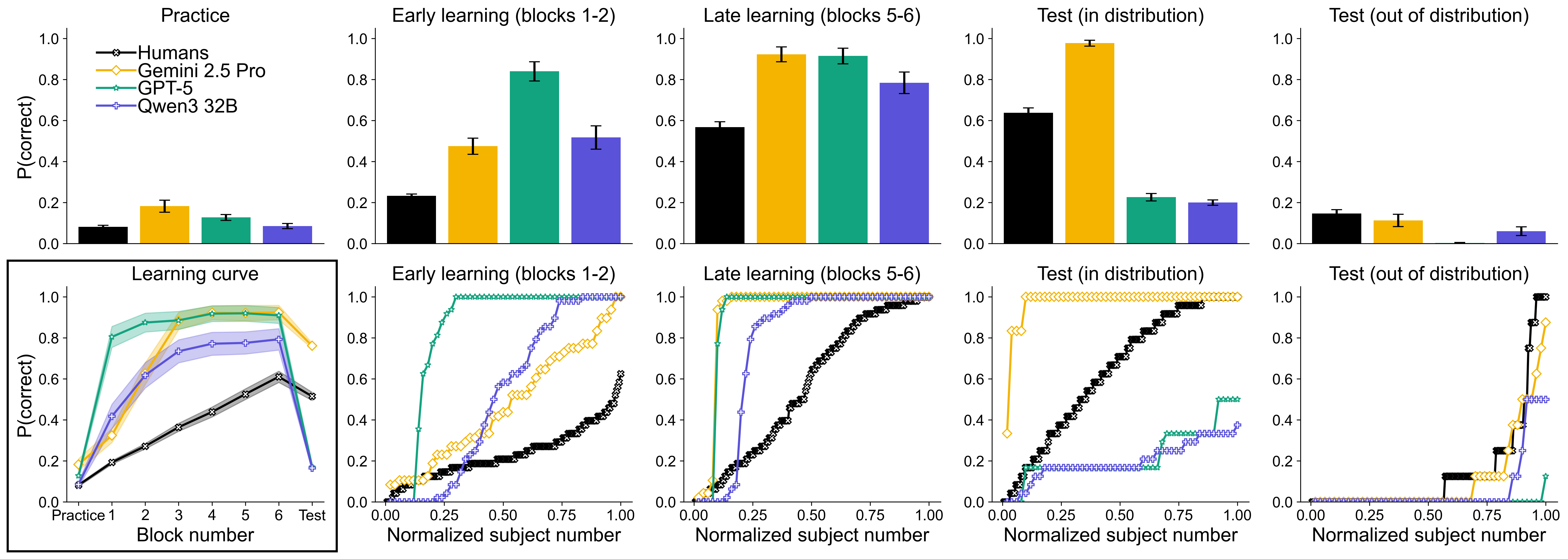}
    \caption{Performance across task phases for humans and models prompted with persona steering.}
    \label{fig:persona_learning}
\end{figure}

\begin{figure}[ht!]
    \centering
    \includegraphics[width=1\linewidth]{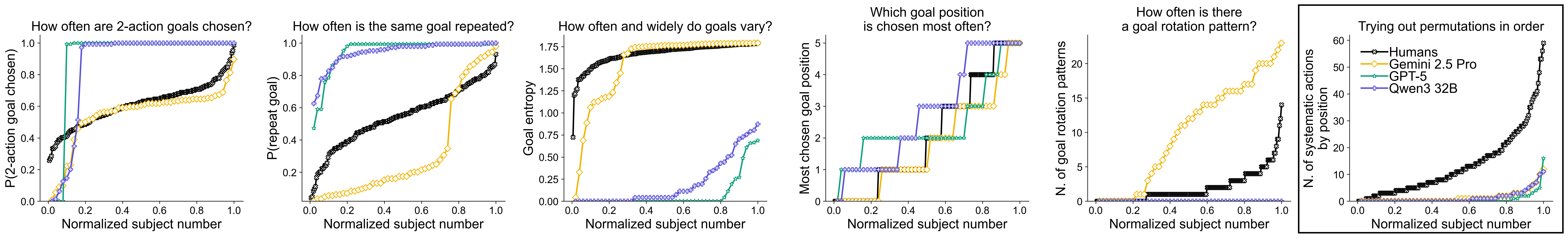}
    \caption{Distributions of goal and action selection behaviors in humans and models prompted with persona steering.}
    \label{fig:persona_goal_action_selection}
\end{figure}

\newpage

\section{Setting the Temperature to Zero} \label{app:temp0}

Below, we report results from our main experimental design after setting the temperature to 0 in models that allowed for this parameter to be customized. Centaur's behavior got significantly more repetitive, and its learning performance -- while similar to humans' on average -- followed a bimodal distribution. Most other findings are consistent with results obtained with a temperature of 1. Note that all variance observed in model simulations is accounted for by the difference in inputs (which followed one of 10 possible configurations of the experiment). 

\begin{figure}[h!]
    \centering
    \includegraphics[width=0.9\linewidth]{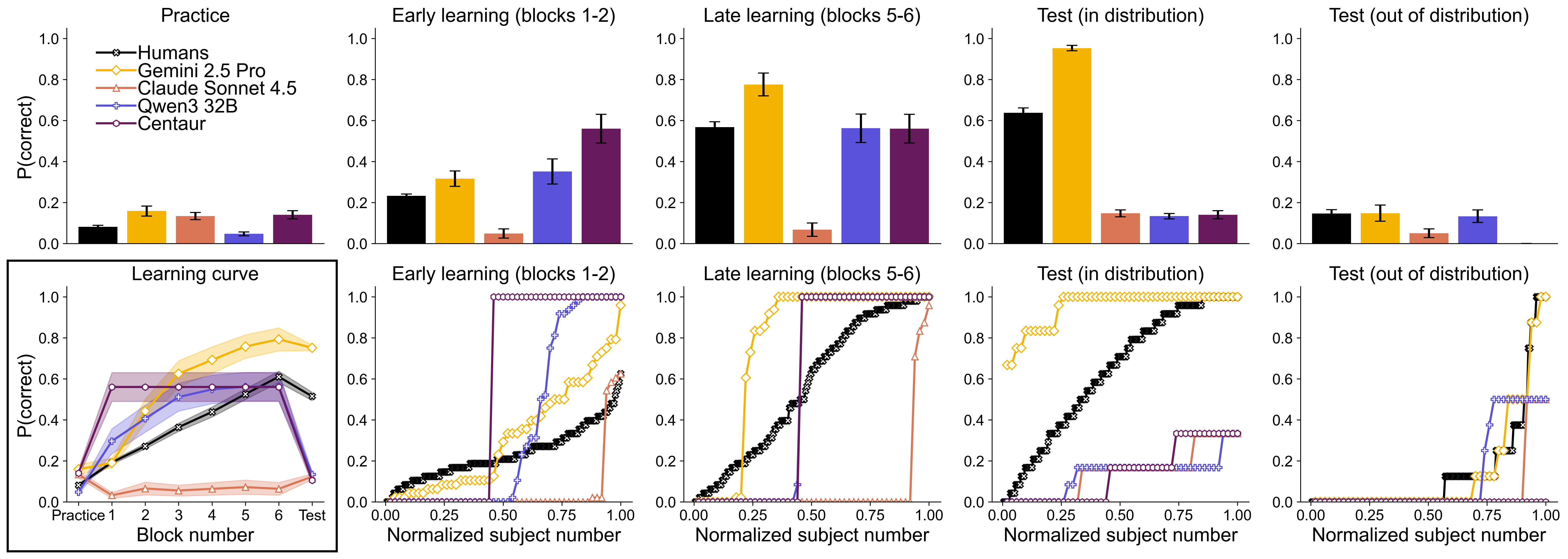}
    \caption{Performance across task phases for humans and models prompted with the temperature parameter set to 0.}
    \label{fig:temp0_learning}
\end{figure}

\begin{figure}[h!]
    \centering
    \includegraphics[width=1\linewidth]{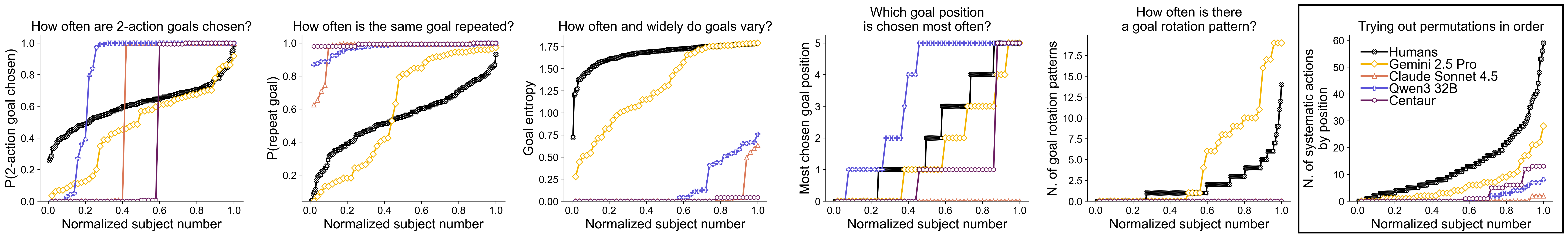}
    \caption{Distributions of goal and action selection behaviors in humans and models  prompted with the temperature parameter set to 0.}
    \label{fig:temp0_goal_action_selection}
\end{figure}
\newpage 

\section{Effects of Semantics}  \label{app:semantics}

\begin{figure}[h!]
    \centering
    \includegraphics[width=1\linewidth]{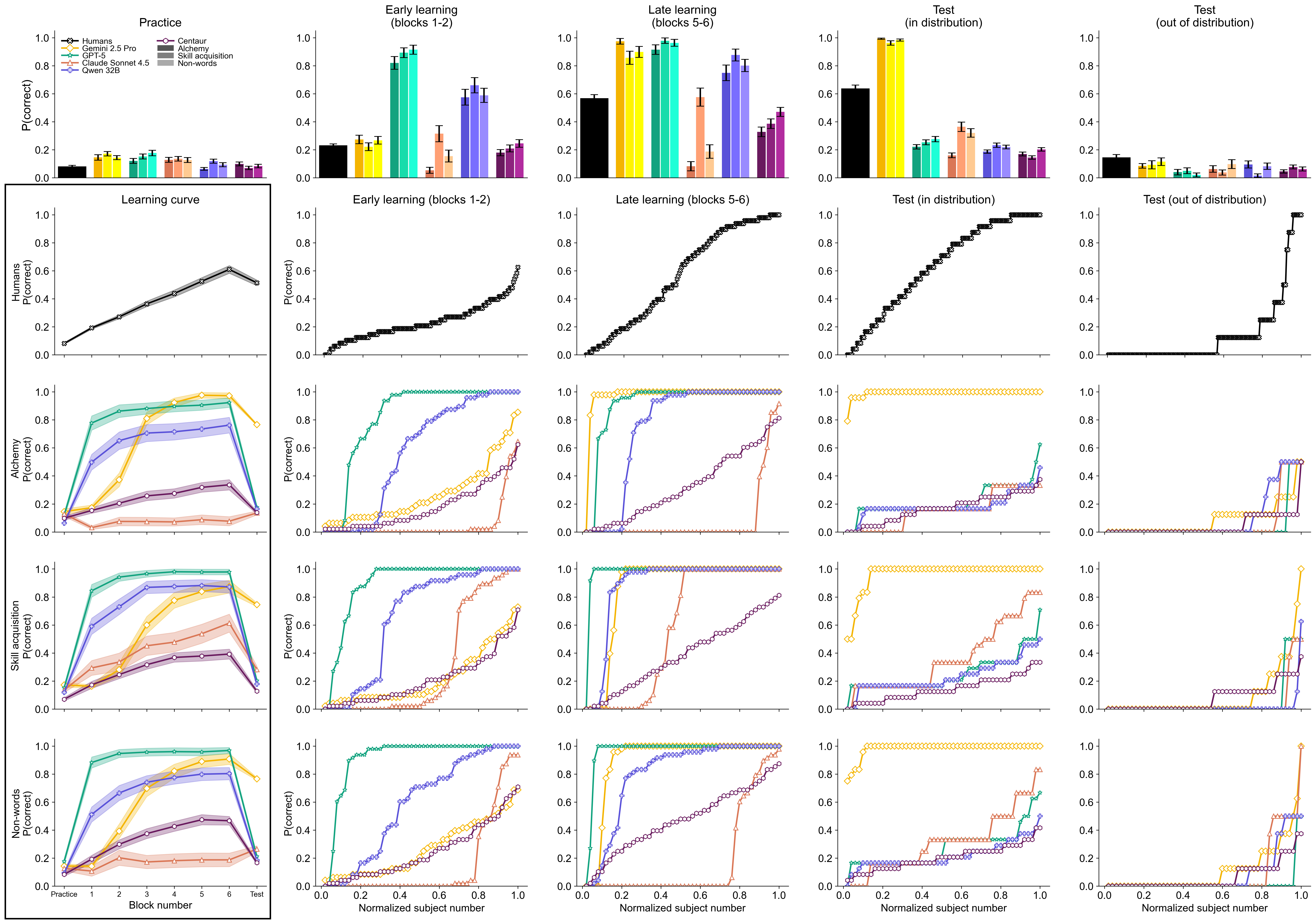}
    \caption{\textbf{Semantics have little impact on task performance across models.} 
    Results show human performance (first row) in the original experiment and LLM goal selection in the alchemy (second row), skill acquisition (third row), and non-words variants of the task (fourth row). 
    Top: average performance in the practice, early learning, late learning, and test blocks across human and LLM task versions, with bar shades indicating the experiment version. 
    Bottom, left-most subplots: learning curves. Bottom, following subplots: sorted individual participant scores, with the x-axis normalized by the number of participants, such that it represents the proportion of participants with a score equal to or lower than the current y. Error bars and shading indicate the S.E.M.}
    \label{fig:semantics_performance}
\end{figure}
\newpage 

\begin{figure}[h!]
    \centering
    \includegraphics[width=1\linewidth]{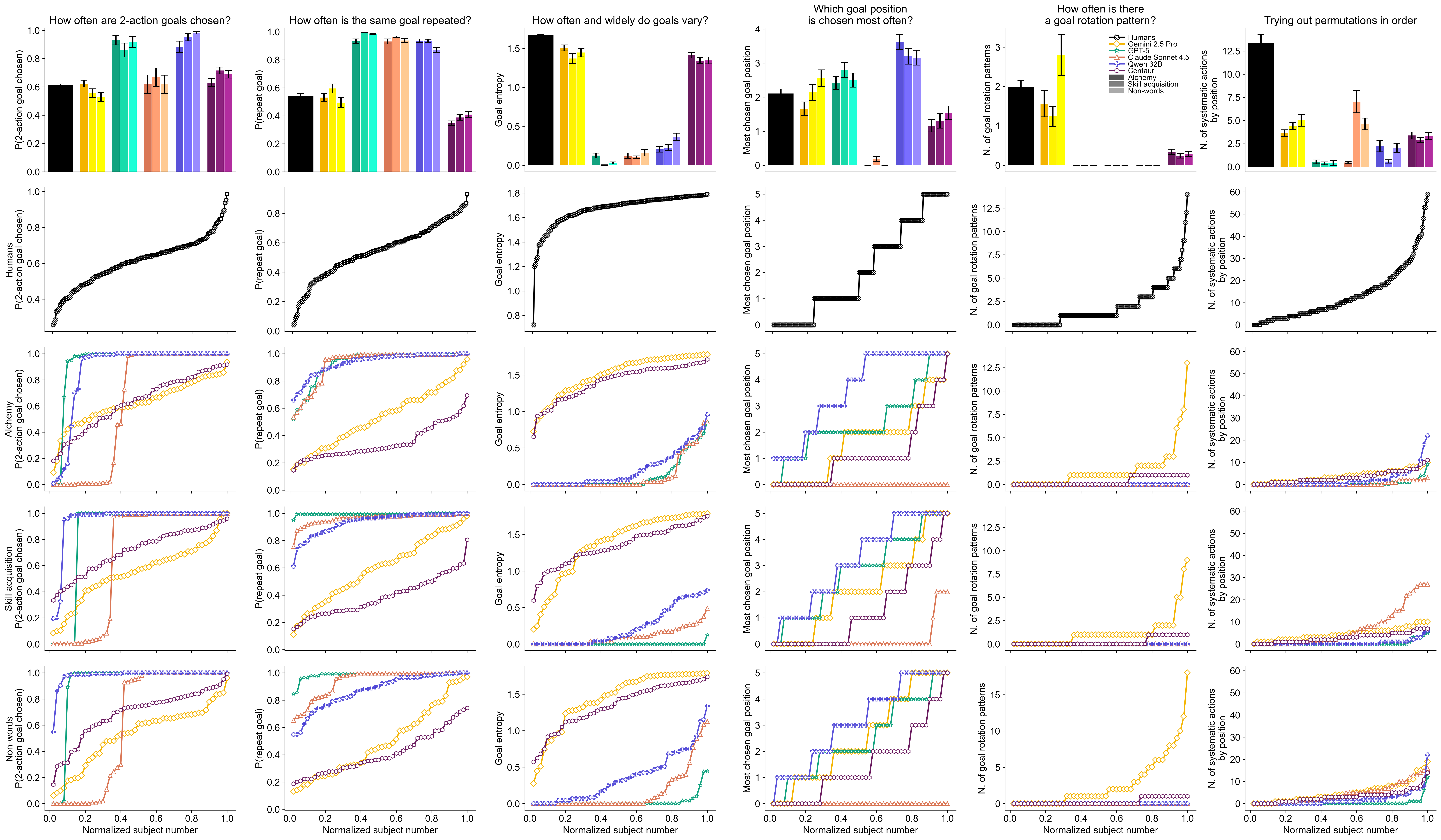}
    \caption{\textbf{Semantics have little impact on goal selection across models.} Results show human goal selection (first row) in the original experiment and LLM goal selection in the alchemy (second row), skill acquisition (third row), and non-words variants of the task (fourth row). Each plot shows the sorted individual scores over the normalized subject number for various aspects of goal (first five subplots) and action selection within repeated goals (right-most subplot). }
    \label{fig:semantics_goal_selection}
\end{figure}
\newpage 

\section{Matching Human and LLM Input Modalities}\label{app:interactive}
In the main experiments discussed so far, LLMs interacted with a text-based environment rather than a visual game (see e.g., \cite{binz2023using, binz2025foundation, coda2024cogbench} for similar approaches). In principle, differences in goal selection between humans and LLMs could be attributed to this implementation difference, which could be tested by matching the input modalities of humans and LLMs.
We note that doing so introduces a confound: any observed divergence between humans and models could then be attributed to differences in visual perception capabilities rather than goal selection propensities. By using a text-based interface for LLMs, we ensure that observed behavioral differences reflect goal selection behavior, not perceptual ones. 
Nonetheless, we conducted an exploratory study with GPT-5.4, whose computer-using agent abilities (\url{https://openai.com/index/computer-using-agent/}) enable it to participate in psychology experiments exactly like a human would, through an online interface. 

To do this, we reconstructed the online platform used by \cite{molinaro2023goal} on a static server and created a URL that the model could access. We set its reasoning to ``medium'' with a liberal cap of 500 interactions per run, and instructed the model to complete the task detailed at the provided URL. Next, the model performed the task exactly as a human participant would, including the full instructions shown in \cite{molinaro2024latent} and training, learning, and testing phases. 
In the results below, we refer to this model as GPT-5.4 CUA (computer-using agent). We compare its behavior to that of humans, and show GPT-5 averages and distributions for visual comparison. 

\begin{figure}[h!]
    \centering
    \includegraphics[width=0.6\linewidth]{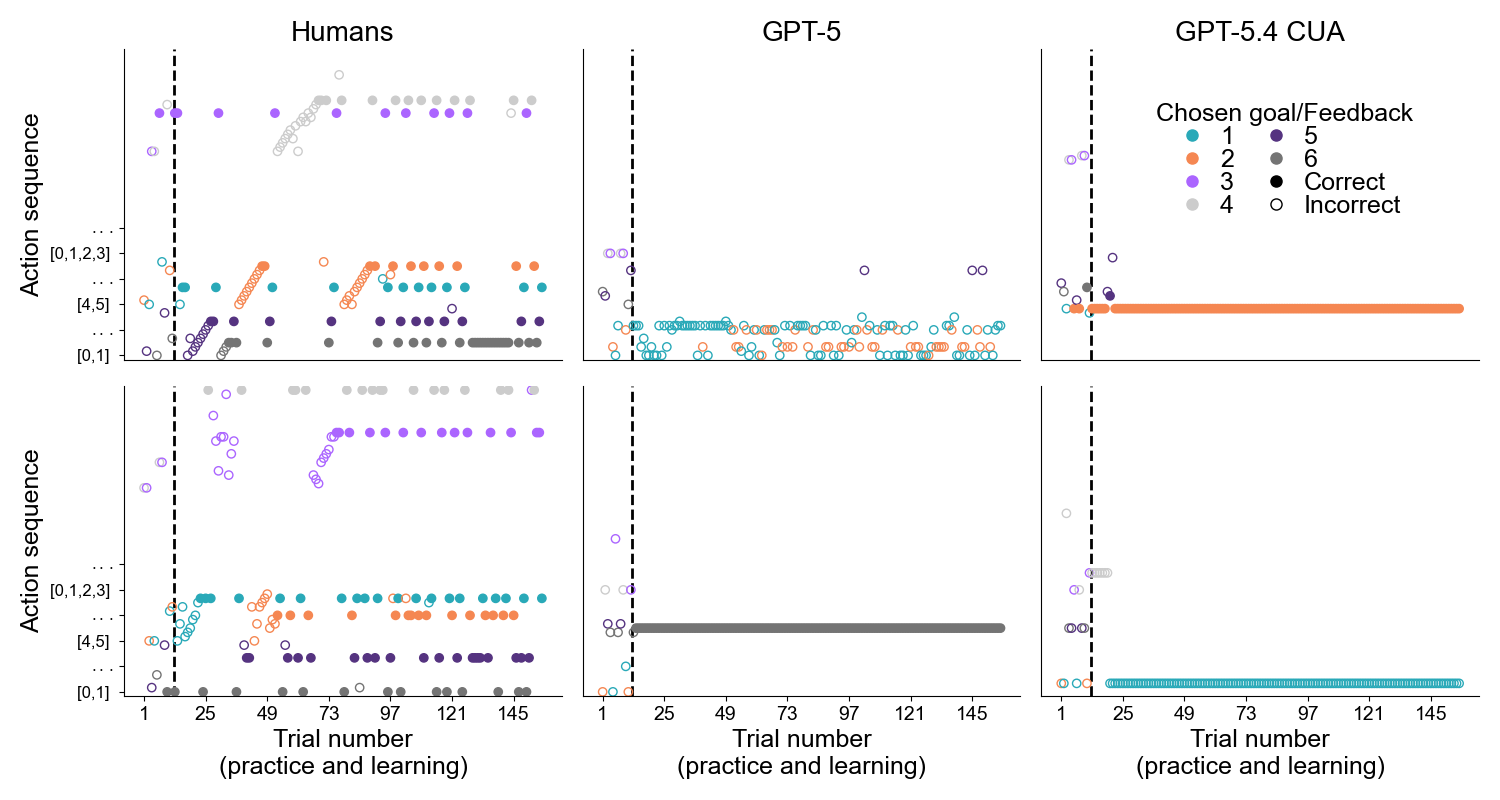}
    \caption{\textbf{Example goal and action selections from humans, GPT-5 (text-based), and GPT-5.4 CUA (interactive online task).}}
    \label{fig:interactive_example_actions}
\end{figure}

\begin{figure}[h!]
    \centering
    \includegraphics[width=0.9\linewidth]{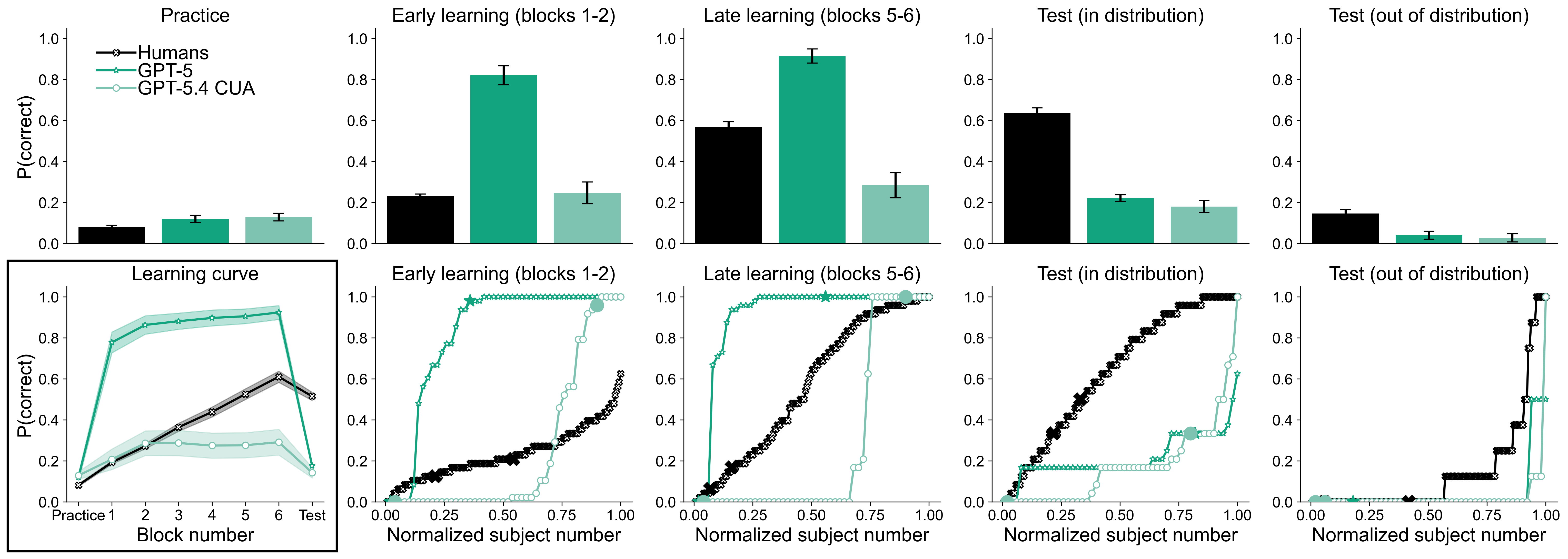}
    \caption{\textbf{Human and GPT-5.4 CUA performance.} GPT-5.4 CUA tends to have poor performance at both learning and testing, with bimodal distributions in the former.}
    \label{fig:interactive_performance}
\end{figure}

\begin{figure}[h!]
    \centering
    \includegraphics[width=1\linewidth]{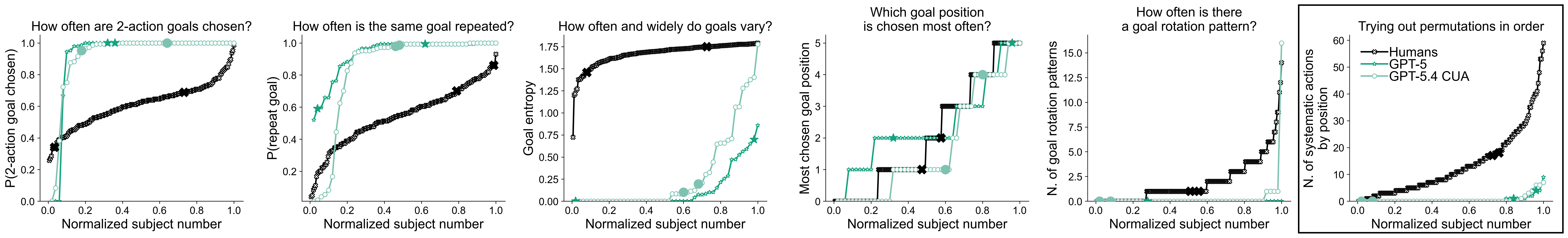}
    \caption{\textbf{Human and GPT-5.4 CUA goal selection.} GPT-5.4 CUA differs from humans on all goal selection metrics tested, except goal position preferences.}
    \label{fig:interactive_summative_metrics}
\end{figure}

When allowed to interact with the task directly via the web interface, GPT-5.4 tended to re-select the same goals and actions, regardless of their correctness -- unlike GPT-5, which favored successful goal-action sequence combinations (Figure \ref{fig:interactive_example_actions}).  
Overall, we find that even when perfectly matching the input modality, the behavior of GPT-5.4 differed significantly from human averages and distributions on all metrics (Tables \ref{tab:stats_descr_base_models_interactive}-\ref{tab:stats_human_vs_llm_base_models_interactive}, both on performance (Figure \ref{fig:interactive_performance}) and goal selection metrics (Figure \ref{fig:interactive_summative_metrics}). The only exception was on goal position preferences, which, similar to GPT-5, did not differ significantly from human averages. 

\begin{table}[h!]
    \centering
    \caption{Descriptive statistics for the interactive experiment.}
    \label{tab:stats_descr_base_models_interactive}
    \scriptsize
    \begin{tabular}{ll}
    \toprule
     & GPT-5.4 CUA \\
    Metric &  \\
    \midrule
    P(corr.) learn & 0.27 ± 0.41 (0.00) \\
    P(corr.) in-distr. & 0.18 ± 0.21 (0.17) \\
    P(corr.) out-distr. & 0.03 ± 0.14 (0.00) \\
    P(2-act goal) & 0.93 ± 0.21 (1.00) \\
    P(repeat goal) & 0.84 ± 0.31 (0.99) \\
    Goal entropy & 0.30 ± 0.47 (0.00) \\
    Pref. goal pos. & 1.76 ± 1.73 (1.00) \\
    N. goal cycles & 0.40 ± 2.24 (0.00) \\
    N. hyp. test. & 0.58 ± 1.64 (0.00) \\
    \bottomrule
    \end{tabular}
\end{table}

\begin{table}[h!]
    \centering
    \caption{Human-AI comparison statistics for the interactive experiment.}
    \label{tab:stats_human_vs_llm_base_models_interactive}
    \scriptsize
    \begin{tabular}{lll}
    \toprule
     & \multicolumn{2}{c}{GPT-5.4 CUA} \\
     \cmidrule(lr){2-3} & Md. & Dist. \\
    Metric &  &  \\
    \midrule
    P(corr.) learn & 2575.5*** & 0.623*** \\
    P(corr.) in-distr. & 1140.5*** & 0.666*** \\
    P(corr.) out-distr. & 2802.0*** & 0.354*** \\
    P(2-act goal) & 8167.0*** & 0.871*** \\
    P(repeat goal) & 7474.5*** & 0.794*** \\
    Goal entropy & 185.0*** & 0.929*** \\
    Pref. goal pos. & 3869.5 & 5.90 \\
    N. goal cycles & 1534.5*** & 0.631*** \\
    N. hyp. test. & 437.5*** & 0.820*** \\
    \bottomrule
    \end{tabular}
\end{table}

\clearpage

\section{Manipulating the Size of the Context Window}\label{app:cw}

\begin{figure}[h!]
    \centering
    \includegraphics[width=1.0\linewidth]{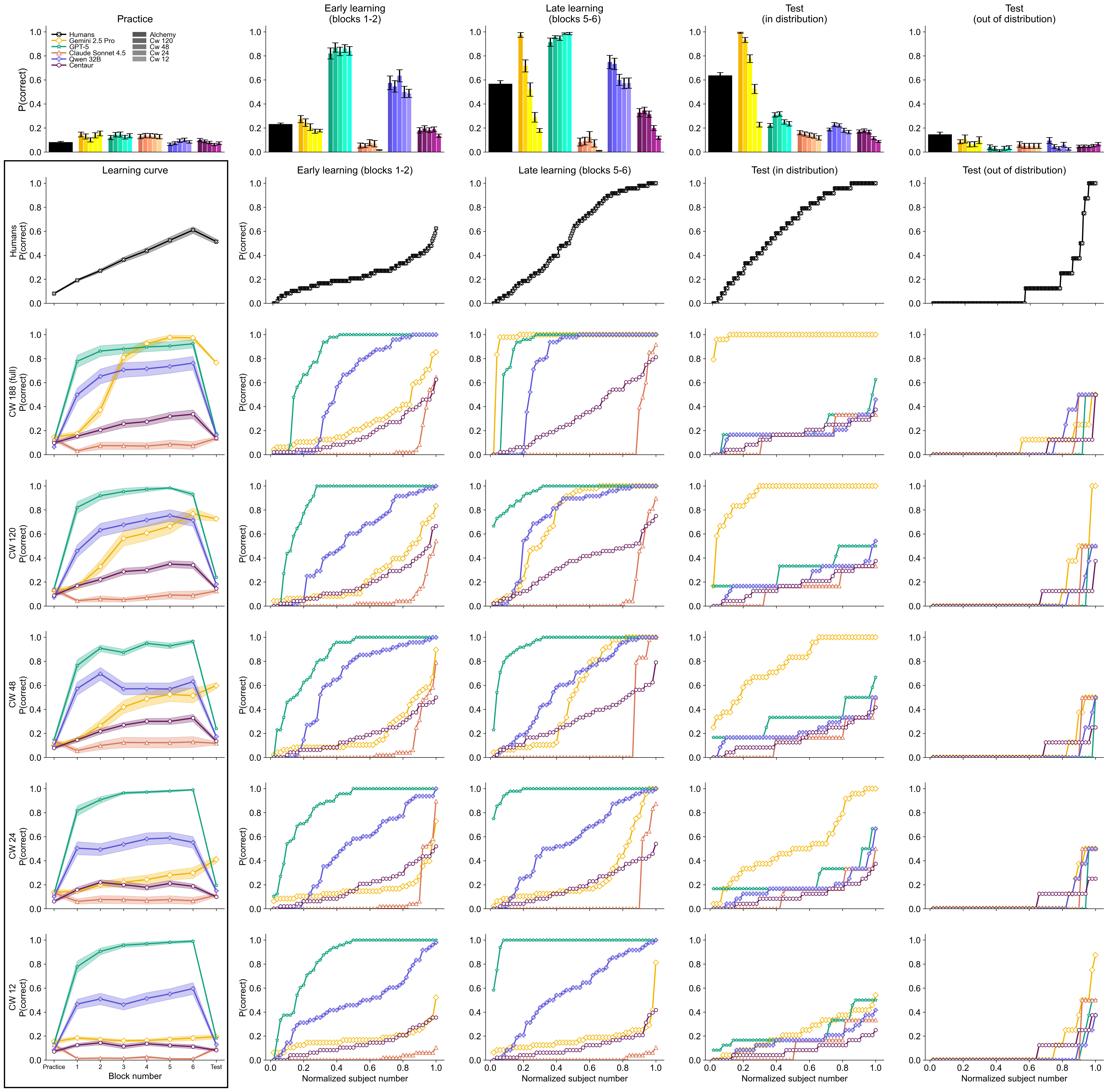}
    \caption{\textbf{LLM goal-contingent performance is negatively impacted by context window size.} Most models show poorer goal-contingent performance with smaller context window sizes.}
    \label{fig:cw_performance}
\end{figure}

LLMs had access to their complete interaction history throughout the experiment, which may explain why they did not need to resort to the same systematic hypothesis-testing as (memory-limited) humans. 
As a first approximation to model memory limitations in LLMs, we ran our main experiment (in the alchemy domain, without additional manipulations) on all models after setting the context window to 12, 24, 48, and 120 out of the 188 total trials. While most models showed systematic effects of context window on goal-dependent choice accuracy in the learning phase, testing phase, or both (Figure \ref{fig:cw_performance}), we saw little to no change in the goal selection patterns of each model family (Figure \ref{fig:cw_goal_selection}). In particular, systematic hypothesis testing was -- if anything -- reduced by a reduced context window, thus strengthening our claims.
Nonetheless, we acknowledge that trimming the context window is a crude approximation for the limited nature of human memory, and we invite additional work that can more accurately capture its features and consequences on goal selection.

\begin{figure}[h!]
    \centering
    \includegraphics[width=1.0\linewidth]{cross_exp_base_models_humans_vs_alchemy_vs_cw120_vs_cw48_vs_cw24_vs_cw12_goal_selection_cdf}
    \caption{\textbf{LLM goal selection remains different from humans' even after trimming the context window size.} In particular, strategic hypothesis testing was not increased by reducing the context window of LLMs to approximate the limited memory capacity of humans.}
    \label{fig:cw_goal_selection}
\end{figure}

\clearpage

\section{Comparing Goal Commitment in Humans and LLMs}\label{app:holton}

To test the applicability of our findings beyond the specific task used in our main analyses, we ran an additional study by re-implementing the task introduced by \cite{holton2024goal}. People are known to persist in goals that would better be abandoned -- a phenomenon also known as the ``sunk cost fallacy'' \cite{arkes1999sunk}, and which the \cite{holton2024goal} task was developed to study. 
In this specific case, we note that imitating human behavior may \textit{not} be desirable, as persistence biases could lead to suboptimal reward accumulation under certain circumstances \cite{aenugu2025building}. However, a failure to match humans in this task undermines the validity of studies replacing human participants with LLMs when assessing behaviors where goal persistence might be relevant. 

In the original experiment, participants completed a computerized game in which they were asked to fill as many ``fishing nets''  as possible by collecting three kinds of marine animals. At the start of each block, participants were given an empty net to fill with a single animal type (crab, octopus, or fish) of their choice. On each trial, participants accepted an offer for different amounts of each animal. Accepting an offer added animals to the fishing net. Choosing the same animal in consecutive trials allowed participants to accumulate offers; choosing a different animal added the current offer to the net but reset any previously made progress. Filling a net resulted in a new block and awarded a point to the participant. Offer values followed a random walk with sudden ``jumps'' which often made it more optimal to switch from one animal type to a new one, even if it meant forgoing current net contents. 

The original study found that people persist with the current goal long after it has become suboptimal. Here, we reproduce the original paradigm to the best of our abilities and adapt it for LLM use, following a similar implementation as explained in the \hyperref[sec:methods]{Methods} section. We ran each model 40 times, 10 for each of the four configurations provided in the original dataset. 
The full prompt used for this study is provided in Appendix \ref{app:prompt_holton}. We refer the reader to \cite{holton2024goal} for details about the task and models used to calculate goal persistence biases. Note that while the original study tested both healthy individuals and patients with brain lesions, we only focus on the healthy individuals from the first experiment.

In comparing humans and LLMs, we focus our analyses on three task-relevant metrics: 
\begin{itemize}
    \item Indifference point (IP): the value of abandonment, calculated through a tree-search model, at which a participant or agent is equally likely to persist or abandon. An optimal agent has an IP of 0. IP values above 0 indicate a tendency to persist with the current goal beyond optimality. IP values below 0 correspond to premature switching behavior. 
    \item Total number of goals completed: the total number of nets filled over the entire task.
    \item The probability of choosing the best alternative after switching from a suboptimal goal: the number of times the best alternative was chosen over all trials in which the participant switched away from the current goal and at least one better offer was available.
\end{itemize}

\begin{figure}[h!]
    \centering
    \includegraphics[width=0.8\linewidth]{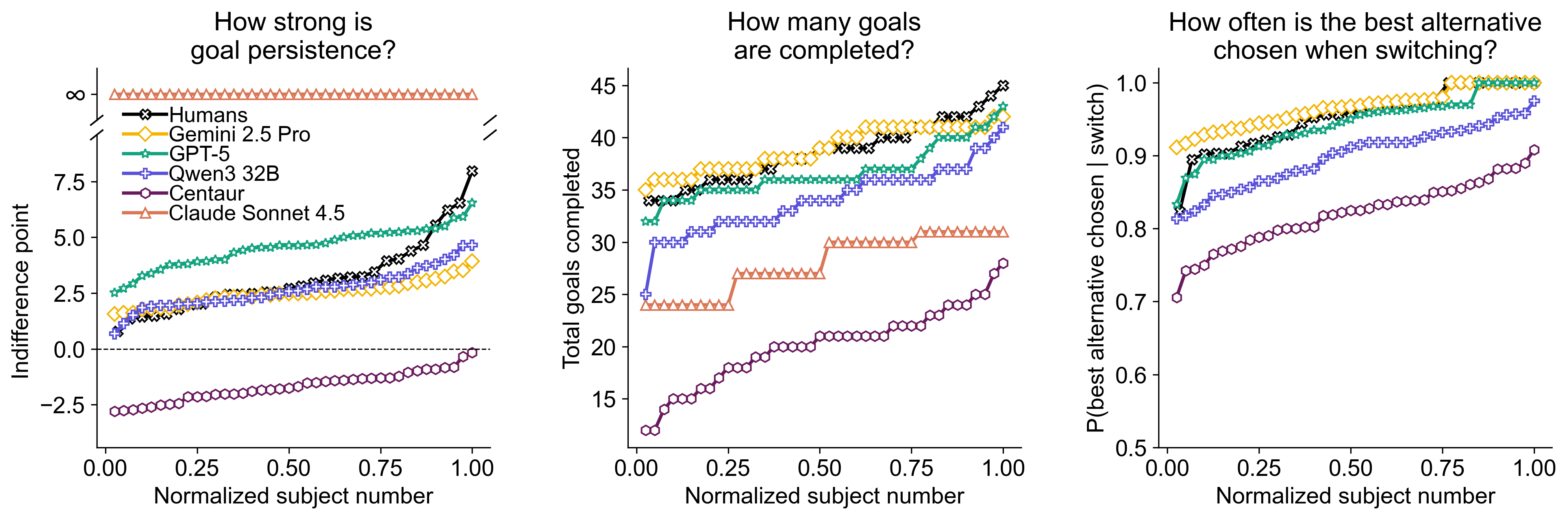}
    \caption{\textbf{Distributions of behavioral scores in the goal commitment task.} Sorted individual scores over the normalized subject number for the goal persistence bias, total number of goals completed over the course of the task, and probability of choosing the best alternative offer when the current goal was suboptimal. Claude Sonnet 4.5 only switched in 10/40 iterations, and only once in each case; therefore, its persistence bias is theoretically infinite; moreover, Claude Sonnet 4.5's probability of choosing the best alternative after switching is impossible to calculate in most cases.}
    \label{fig:goal_commitment_metrics}
\end{figure}

Different models matched human signatures of goal commitment to varying degrees (Figure \ref{fig:goal_commitment_metrics}; Tables \ref{tab:stats_descr_goal_commitment}-\ref{tab:stats_human_vs_llm_goal_commitment}).
Both Qwen3 32B and Gemini 2.5 Pro had similar persistence biases to humans, while other models ranged across extremes. 
On one end of the spectrum, Claude Sonnet 4.5 had a theoretically infinite indifference point, as it only abandoned the initially chosen goals in 10/40 runs, and only once in each simulation -- practically fixating on one goal no matter the external conditions. This could not be due to the model fixating on the first listed option, as the offer order was randomized on each trial. 
GPT-5 also showed more persistent behavior than human participants, but not as severely. 
On the other end of the spectrum, Centaur showed a \textit{negative} persistence bias, switching more often than required by the task -- consistent with the unstructured behavior we saw in our original experiment (Figure \ref{fig:example_actions}).

Even with a similar persistence bias, Qwen3 32B completed fewer goals than humans, potentially because it was worse than humans at selecting the best option after abandoning. 
By contrast, Gemini 2.5 Pro had near-optimal behavior post abandonment, resulting in a total number of completed goals above the human average, although not significantly so. 
GPT-5 chose the best option after a switch as efficiently as humans, but because of its stronger persistence bias, it did not manage to fill as many nets, on average.
Claude Sonnet 4.5 and Centaur completed fewer goals and chose the best alternative less efficiently than humans after a switch. 

Together, these results show that most models differ from human averages and distributions of goal commitment. 
Moreover, seemingly overlapping biases -- seen in Qwen3 32B -- appear to stem from a general suboptimality in behavior, rather than a specific characteristic shared with humans.  
The surprising result that Centaur switched more often than normal is consistent with the erratic behavior observed in our main experiment (Figure \ref{fig:example_actions}), and could be partly due to the fact that it often failed to output one of the available options, requiring resampling (other models required minimal or no resampling).
Interestingly, Gemini 2.5 Pro behaved close to humans on all three metrics we selected, in line with the finding that LLMs can inherit human-like biases from human data \cite{acerbi2023large, lampinen2024language}. 
However, because data from \cite{holton2024goal} was publicly shared, we cannot exclude the possibility that the model was directly trained on it, which could also explain its ability to match human behavior.

Together, these findings show that human-LLM alignment in goal selection without specialized inference techniques is the exception, not the rule. 

\begin{table}[h!]
\centering
\caption{Descriptive statistics for the goal commitment task. IP = indifference point; N. goals = total number of goals completed; P(best|switch) = probability of choosing the best alternative after switching from a suboptimal goal. Claude Sonnet 4.5 did not switch often enough for IP and P(best) to be calculated.}
\label{tab:stats_descr_goal_commitment}
\scriptsize
\begin{tabular}{lllllll}
\toprule
Model & Humans & Gemini 2.5 Pro & GPT-5 & Claude Sonnet 4.5 & Qwen3 32B & Centaur \\
Metric &  &  &  &  &  &  \\
\midrule
IP & 3.17 ± 1.63 (2.76) & 2.49 ± 0.53 (2.49) & 4.55 ± 0.88 (4.63) & --- & 2.67 ± 0.86 (2.60) & -1.68 ± 0.65 (-1.72) \\
N. goals & 38.63 ± 2.97 (39.00) & 38.95 ± 2.07 (39.00) & 36.67 ± 2.52 (36.00) & 28.00 ± 2.74 (28.50) & 34.08 ± 3.19 (34.00) & 20.07 ± 3.70 (21.00) \\
P(best) & 0.95 ± 0.04 (0.96) & 0.97 ± 0.03 (0.97) & 0.94 ± 0.04 (0.95) & --- & 0.90 ± 0.04 (0.91) & 0.82 ± 0.04 (0.82) \\
\bottomrule
\end{tabular}
\end{table}

\begin{table}[h!]
\centering
\caption{Human-AI comparison statistics for the goal commitment task.}
\label{tab:stats_human_vs_llm_goal_commitment}
\scriptsize
\begin{tabular}{lllllllllll}
\toprule
 & \multicolumn{2}{c}{Gemini 2.5 Pro} & \multicolumn{2}{c}{GPT-5} & \multicolumn{2}{c}{Claude Sonnet 4.5} & \multicolumn{2}{c}{Qwen3 32B} & \multicolumn{2}{c}{Centaur} \\
 \cmidrule(lr){2-3} \cmidrule(lr){4-5} \cmidrule(lr){6-7} \cmidrule(lr){8-9} \cmidrule(lr){10-11}
 & Md. & Dist. & Md. & Dist. & Md. & Dist. & Md. & Dist. & Md. & Dist. \\
Metric &  &  &  &  &  &  &  &  &  &  \\
\midrule
IP & 468.0 & 0.308 & 959.0*** & 0.625*** & --- & --- & 524.0 & 0.167 & 0.0*** & 1.000*** \\
N. goals & 651.5 & 0.150 & 381.0** & 0.383** & 0.0*** & 1.000*** & 188.0*** & 0.533*** & 0.0*** & 1.000*** \\
P(best) & 693.0 & 0.233 & 511.5 & 0.192 & --- & --- & 231.0*** & 0.542*** & 25.0*** & 0.942*** \\
\bottomrule
\end{tabular}
\end{table}

\clearpage
\section{Prompts} \label{app:prompt}
Below, we report information about the prompts used in our study. Additional empty lines were omitted to save space.

\subsection{Main Study} \label{app:prompt_main}
Prompts for each trial started with an introduction to the game:
\newline
\indentedsay{You are participating in an alchemy game where you create potions by combining ingredients.
\newline
HOW THE GAME WORKS:\newline
• In each trial, you will select ingredients to create a specific potion.\newline
• Each potion requires either 2 or 4 specific ingredients, combined in a particular sequence.\newline
• You will select the required ingredients one by one, in order.\newline
• Each ingredient can only be used once per attempt.\newline
• After selecting all ingredients, you will see whether the flask fills with the potion or remains empty.\newline
IMPORTANT RULES:\newline
• Every potion has one correct recipe (specific ingredients in a specific order).\newline
• The recipe for each potion stays the same throughout the entire game.\newline
• Some potions can be created using multiple methods - either by combining basic ingredients directly, or by using other completed potions as ingredients.\newline
• If your ingredient sequence matches the correct recipe, the flask will fill with the potion.\newline
 If your sequence doesn't match, the flask will remain empty.}
\newline
This was followed by additional information, which was dependent on the specific configuration of the task (pseudo-randomly assigned). Below is an example based on one task configuration:
\newline
\indentedsay{AVAILABLE POTIONS:\newline 
Potion 0:\newline   
• Color: green\newline   
• Ingredients needed: 4\newline   
• Available ingredients: mushrooms, butterfly, horseshoe, frog\newline 
Potion 1:\newline   
• Color: yellow\newline   
• Ingredients needed: 2\newline   
• Available ingredients: mushrooms, butterfly, horseshoe, frog\newline  
Potion 2:\newline   
• Color: pink\newline   
• Ingredients needed: 2\newline   
• Available ingredients: red potion, orange potion, yellow potion, purple potion\newline Potion 3:\newline   
• Color: pink\newline   
• Ingredients needed: 4\newline   
• Available ingredients: mushrooms, butterfly, horseshoe, frog\newline  
Potion 4:\newline   
• Color: purple\newline   
• Ingredients needed: 2\newline   
• Available ingredients: mushrooms, butterfly, horseshoe, frog\newline  
Potion 5:\newline   
• Color: green\newline   
• Ingredients needed: 2\newline   
• Available ingredients: red potion, orange potion, yellow potion, purple potion}
\newline
Starting from the second trial, information about the preceding trials was also included, which differed based on the task phase and the correctness of the ingredient sequence. E.g., in the practice phase:
\newline
\indentedsay{PREVIOUS EXPERIMENTS:
\newline Trial 1: [Training] You were assigned potion 0 (green) and chose ingredients ['horseshoe', 'frog', 'mushrooms', 'butterfly'] - the flask remained empty.}
\newline
In the learning phase:
\newline
\indentedsay{PREVIOUS EXPERIMENTS:
\newline Trial 1: [Training] You were assigned potion 0 (green) and chose ingredients ['horseshoe', 'frog', 'mushrooms', 'butterfly'] - the flask remained empty.
\newline [Information about trials 2-12 omitted for brevity]
\newline Trial 13: [Learning] You chose potion 0 (green) and chose ingredients ['mushrooms', 'butterfly', 'horseshoe', 'frog'] - the flask filled with the potion.}
\newline
In the test phase:
\newline
\indentedsay{PREVIOUS EXPERIMENTS:
\newline 
[Information about trials 2-156 omitted for brevity]
\newline Trial 157: [Testing] You were assigned potion 0 (green) and chose ingredients ['mushrooms', 'butterfly', 'horseshoe', 'frog'] - no feedback given.}
\newline
In the practice and test phases, where goals were forced, the following text was used, with the specific potion varying on a trial-by-trial basis, e.g.:
\newline
\indentedsay{You have been assigned to create potion 0 (green).}
\newline
In the learning phase, models were prompted to pick a goal themselves, e.g.:
\newline
\indentedsay{TRIAL 13:\newline
Q: Which potion would you like to create? \newline
Available options: [0, 1, 2, 3, 4, 5].\newline
Please respond with the number of your chosen potion:\newline
A: }
\newline
Next, models were prompted to select ingredients until all slots were filled, e.g.:
\newline
\indentedsay{TRIAL 13: You have chosen to create potion 0 (green).
\newline
This potion requires 4 ingredients in a specific sequence. \newline
INGREDIENT SELECTION:\newline
• Ingredients selected so far: ['mushrooms', 'butterfly']\newline
• Remaining available ingredients: ['horseshoe', 'frog']\newline
Q: Select ingredient 3 of 4:\newline
Please respond with the exact name of your chosen ingredient:
}
\newline
Then, model choices and (for practice and learning phases) feedback were added to the next trial's input as exemplified above.

\subsection{Chain of Thought} \label{app:prompt_cot}
Following \cite{coda2024cogbench}, we created a separate prompt to induce chain-of-thought reasoning in a subset of the models.
This was achieved by appending the following sentence to the prompt, replacing instructions on how to answer concisely:
``First break down the problem into smaller steps and reason through each step logically in a maximum of 100 words before giving your final answer in the format 'Final answer: <your choice>'. 
It is very important that you always answer in the right format even if you have no idea or you believe there is not enough information.
A: Let's think step by step: 1. ''.
CoT traces were not appended to the trial-by-trial history.

\subsection{Persona Steering} \label{app:prompt_persona}
In a separate version of the experiment, we prompted models with the following ``persona'' description, prepended to the rest of the prompt:
``You are a university student participating in a psychology study to earn course credit.
You don't know what the study is really about, but you want to do your best and answer honestly, as a typical participant would.
Please respond naturally to all instructions and questions as if you were an undergraduate student taking part in a real experiment.''

\subsection{Skill Acquisition Setup} \label{app:prompt_skill_acquisition}
In a separate version of the task, we framed the task as an ``acrobatics game'' with the following instructions:
\newline
\indentedsay{You are participating in an acrobatics game where you perform specific moves.\newline
HOW THE GAME WORKS:\newline
• In each trial, you will select actions to create a specific acrobatic move.\newline
• Each move requires either 2 or 4 specific actions, combined in a particular sequence.\newline
• You will select the required actions one by one, in order.\newline
• Each action can only be used once per attempt.\newline
• After selecting all actions, you will see whether the acrobatic routine is successful or not.\newline
IMPORTANT RULES:\newline
• Every move has one correct sequence (specific actions in a specific order).\newline
• The sequence for each move stays the same throughout the entire game.\newline
• Some moves can be created in multiple ways - either by combining basic actions directly, or by using other completed moves as actions.\newline
• If your action sequence matches the correct sequence, the routine will be successful.\newline
• If your sequence doesn't match, the routine will fail.}
\newline
Everywhere else, we replaced the word ``potion'' with ``move'', and the word ``ingredient'' with ``action''.
Specific potions were replaced with the names of fictional athletes (a pseudo-random subset of Vaskov, Miretti, Oduya, Pelran, Cahill, Kweon, Szabó, Tale, and Ping). Specific ingredients were replaced with the actions ``arch'', ``tuck'', ``straddle'', ``jump'', and already completed ``moves'' (similar to pre-made potions in the original task). The structure of the task itself was otherwise identical to the original.

\subsection{Non-Words Setup} \label{app:prompt_nonwords}
To present a semantics-agnostic version of the environment to LLMs, we replaced all key elements of the game with ``non-words'' borrowed from previous studies on human category learning or developmental psychology (e.g., \cite{markman1984children, gopnik2000detecting}).
We replaced all instances of the word ``potion'' with ``dax'', and all instances of the word ``ingredient'' with ``blicket''. The potion elements themselves from the original game were replaced with a subset of the following non-words: neem, fep, wub, zib, gorp, gazzer, mido, jop, and fop. Basic ingredients were replaced with the non-words lif, toma, modi, koba.
The remaining instructions were presented in plain English as in the original version.

\subsection{Goal Commitment Task} \label{app:prompt_holton}

Prompts for each trial started with an introduction to the game:
\newline
\indentedsay{
    You are participating in a deep-sea fishing game where you fill nets with seafood to earn points.
    \newline
    HOW THE GAME WORKS:\newline
    • Each round, you are offered quantities of three types of seafood: crab, octopus, and fish.\newline
    • You choose one type to add to your net.\newline
    • Your net can only hold ONE type of seafood at a time.
    • If you choose the SAME type already in your net, the offered quantity is added to your net.\newline
    • If you choose a DIFFERENT type, ALL seafood currently in your net is LOST, and the net starts fresh with only the newly chosen quantity.\newline
    • Offers are usually positive, but can sometimes be negative. Choosing a negative offer subtracts that amount from your net. Your net cannot go below zero.\newline
    • Offers for each type change gradually from round to round, but can sometimes jump dramatically.\newline
    • When your net reaches its capacity, you score a point and begin filling a new net.\newline
    • Your goal is to score as many points as possible within the limited number of rounds.\newline
    CURRENT STATE:\newline
    • Net: empty (capacity: 33)\newline
    • Points scored: 0\newline
    • Rounds remaining: 300} 
\newline 
Then, an offer was generated, based on the pseudo-randomized schedules described in \cite{holton2024goal}. As in the original study, the order in which offers were presented varied on a trial-by-trial basis. Offers were rounded to 4 decimal points. For example, the first offer set could be:
\newline
\indentedsay{
    CURRENT OFFERS:\newline
    • Crab: 5.2989\newline
    • Octopus: 5.2569\newline
    • Fish: 4.4532\newline 
    Round 1 of 300. Which seafood do you choose: Crab, Octopus, Fish?
    \newline 
    Your choice:
}
\newline 
At this point, the model's choice was recorded, added to the current net, and appended to the model's interaction history. The number of rounds left kept decrementing. For instance, the second trial's prompt could include the following, before offers were presented: 
\newline
\indentedsay{
    PREVIOUS ROUNDS:\newline
    Round 1: You chose crab (+5.2989). Net: 5.2989/33 crab.
    \newline
    CURRENT STATE:\newline
    • Net: 5.2989 crab (capacity: 33)\newline
    • Points scored: 0\newline
    • Rounds remaining: 299}
\newline
After a net was filled (i.e., accumulated offers reached or exceeded capacity), a point was added to the current score, and the net was reset. 



\end{document}